\pdfoutput=1

\documentclass[11pt]{article}

\usepackage[final]{acl}

\usepackage{times}
\usepackage{latexsym}

\usepackage[T1]{fontenc}

\usepackage[utf8]{inputenc}

\usepackage{microtype}

\usepackage{inconsolata}

\usepackage{paralist}
\usepackage{subcaption}
\usepackage{amssymb}
\usepackage{threeparttable}
\usepackage{fancyhdr}
\usepackage{multirow}
\usepackage{tikz}

\newcommand{\perfsim}{\ensuremath{s_\mathit{perf}}}
\newcommand{\semanticsim}{\ensuremath{s_\mathit{sem}}}
\newcommand{\perfmatrix}{\ensuremath{Q}}
\newcommand{\perfmatrixvector}{\ensuremath{\mathbf{q}}}
\newcommand{\similaritymatrix}{\ensuremath{T}}

\title{Examining the robustness of LLM evaluation \\ to the distributional assumptions of benchmarks}
 
\author{Melissa Ailem\footnotemark[2] \and Katerina Marazopoulou\footnotemark[2] \and Charlotte Siska\footnotemark[2] \and James Bono \\
        Microsoft\\
        \texttt{\{melissaailem, aikaterinim, csiska, jabono\}@microsoft.com}
}

\begin{document}

\maketitle
\begingroup
\def\thefootnote{\textdagger}\footnotetext{These authors contributed equally to this work.}
\endgroup

\begin{abstract}
Benchmarks have emerged as the central approach for evaluating Large Language Models (LLMs). 
The research community often relies on a model's average performance across the test prompts of a benchmark to evaluate the model's performance. 
This is consistent with the assumption that the test prompts within a benchmark represent a random sample from a real-world distribution of interest. 
We note that this is generally not the case;  
instead, we hold that the distribution of interest varies according to the specific use case. 
We find that 
\begin{inparaenum}[(1)]
    \item the correlation in model performance across test prompts is non-random, 
    \item accounting for correlations across test prompts can change model rankings on major benchmarks, 
    \item explanatory factors for these correlations include semantic similarity and common LLM failure points.
\end{inparaenum}

\end{abstract}

\section{Introduction}

Since the introduction of the Transformer architecture~\citep{vaswani2017attention}, Large Language Models (LLMs) have progressed into sophisticated systems with an outstanding ability to comprehend and generate text that mimic human language. 
Notable models in this domain include ChatGPT\footnote{New chat: \url{https://chat.openai.com/}}, utilizing the GPT-3.5-TURBO or GPT-4 architectures\footnote{Models - OpenAI API: \url{https://platform.openai.com/docs/models/}}, LLaMA~\citep{touvron2023llama}, ChatGLM~\citep{zeng2023glm}, Alpaca~\citep{taori2023alpaca}, and Falcon~\citep{penedo2023refinedweb}. 

\begin{figure}
    \captionsetup[subfigure]{position=b}
    \centering
    \subcaptionbox{\label{subfig:unweighted} All prompts contribute equally during evaluation.}
    {\includegraphics[width=.9\columnwidth]{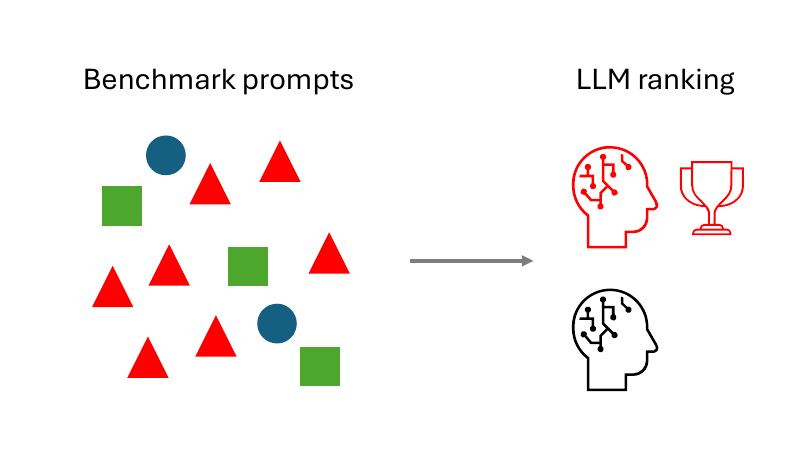}}
    \subcaptionbox{\label{subfig:weighted} Prompts are weighted during evaluation.}
    {\includegraphics[width=.9\columnwidth]{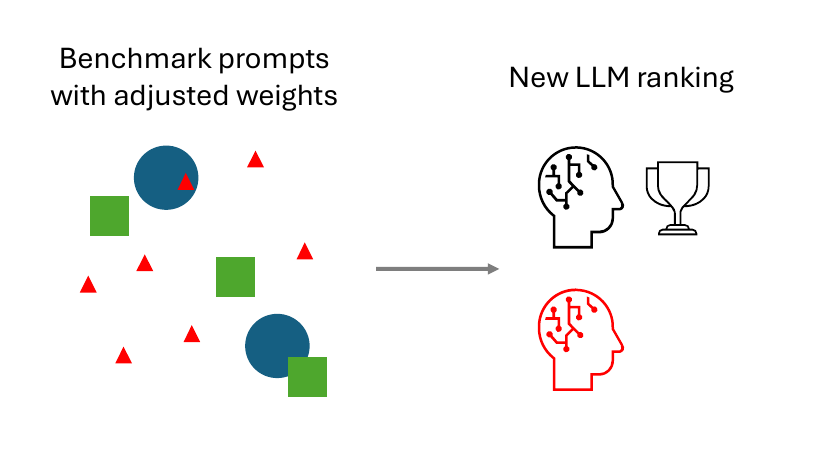}}
\caption{Illustrative example showcasing how different distributional assumptions of benchmarks affect model rankings.
Consider a benchmark containing prompts reflecting three different tasks:
math (red triangles), code generation (blue circles), and text generation (green squares).
In Figure~\ref{subfig:unweighted}, each benchmark prompt contributes equally to the model evaluation.
In contrast, Figure~\ref{subfig:weighted} accounts for correlations between prompts and the weights of the prompts are adjusted accordingly during evaluation.
In scenario~\ref{subfig:unweighted}, the red LLM ranks highest because it excels in math, and the benchmark is biased towards math tasks (7 out of 12 prompts are math-related). 
When considering different weights in scenario~\ref{subfig:weighted}, we observe a different ranking outcome.}
\label{fig:explanatory_figure}
\end{figure}

Due to their effectiveness, LLMs are becoming very popular in both academia and industry, making their evaluation crucial. 
However, this effectiveness comes at the cost of increased complexity, which makes their evaluation very challenging.
Although prior research has introduced benchmarks for different tasks along with evaluation measures, these assessments often overlook potential biases. 
When a benchmark includes multiple prompts with similar characteristics, it can increase or decrease the average performance of a model, so model comparisons can become brittle with respect to benchmark composition (see Figure~\ref{fig:explanatory_figure} for an illustrative example).
In this work, we show that the inherent connections between the prompts in current benchmarks impact the models' performance and their relative rankings.

The standard approach for evaluation on a benchmark is to
\begin{inparaenum}[(i)]
\item obtain model responses for each prompt in the benchmark, 
\item compute the performance metrics for each response,
\item\label{item:aggregation} aggregate (usually average) the performance metrics 
to obtain a single performance metric over the benchmark, and
\item compare models by comparing their aggregate performance. 
\end{inparaenum}

When aggregating performance metrics in step~\ref{item:aggregation} above, each prompt is generally weighted equally~\citep{yang2023large, pena2023leveraging}. 
However, 
using equal weights reflects the assumption that prompts in the benchmark are ``equal'', in the sense that prompts are representative samples of a target distribution of interest. 
In the case of LLMs, the notion of a target distribution (i.e., the distribution of all possible prompts for a given use case) is usually not well-defined. 
For example, different Natural Language Inference (NLI) applications may have very different target distributions, and we should not expect a single benchmark to capture every one. 
Therefore, one must ask: \emph{What distribution do the prompts in the benchmark represent?}
Would considering different distributions fundamentally change model comparisons?
In this work, we present a novel approach to assess the robustness and adequacy of benchmarks used in evaluating LLMs, by analyzing the performance of multiple LLMs on a set of four major benchmarks.

Our key contributions are outlined below:
\begin{asparaenum}
\item 
For each considered benchmark, we observe that the correlation of model performance across prompts is significant (p-value < 0.05). 
This demonstrates the existence of relationships between prompts within the investigated benchmarks. 


\item 
We explore the robustness of model comparisons to different distributional assumptions based on correlation structure, and we observe shifts in performance as large as 10\% and rank changes as large as 5 (out of 14 models).

\item 
We provide a characterization of performance over the distribution of all possible prompt weights. 
This constitutes a robustness check that can be incorporated in comparative studies.

\item We show that model performance similarity across prompts can be explained by semantic similarity, but it is most likely derived by common failure points of the LLM.



\end{asparaenum}

\section{Related work}

Evaluating the performance of LLMs has become a critical area of research, drawing significant attention in recent years.
Comprehensive surveys of LLM evaluation and benchmark quality can be found in~\citet{chang2023survey, guo2023evaluating, perlitz2023}, and~\citet{liang2022holistic}.

When assessing the quality of LLMs, the robustness aspect is  becoming of increasing importance~\citep{wang-etal-2022-measure, goel-etal-2021-robustness}. 
Robustness investigates the stability of a model when confronted with unforeseen prompts. 
Robustness research can be divided into four main lines of work~\citep{li_survey_2023}: 
\begin{inparaenum}[(i)]
\item robustness under distribution shift~\citep{ijcai2021p628, yang2023glue}, 
\item robustness to adversarial input~\citep{zhu2023promptbench,wang2023robustness}, 
\item robustness to prompt formats, including instruction templates~\citep{mizrahi2023, voronov2023, weber2023, sclar2023}, and
\item robustness to dataset bias~\citep{gururangan-etal-2018-annotation, le2020adversarial, niven-kao-2019-probing}.
\end{inparaenum}
Our work falls into the latter category. 

Reducing bias on benchmarks is a long-standing area of research spanning many diverse fields.
Applications range from weighing survey responses to match a target population~\cite{DeBell2018}, to accounting for language biases in visual question-answering~\citep{goyal2017making}.
In the context of NLI, researchers have looked into improving the quality of prompts in order to mitigate certain types of biases. 
Work in this area has focused on determining the quality of prompts by generating optimal prompts~\citep{pryzant-etal-2023-automatic, deng-etal-2022-rlprompt} or by clustering prompts based on semantic similarity ~\citep{kuhn_semantic_2023}. 
Additionally, researchers have investigated data leakage between benchmarks and LLM training data~\citep{zhou_dont_2023, oren_proving_2023}.

Limited research has been conducted to study inherent biases in LLM benchmarks. 
Among existing works, ~\citet{gururangan-etal-2018-annotation} and~\citet{niven-kao-2019-probing} have shown that models leverage spurious statistical relationships in the benchmark datasets and, thus, their performance on the benchmarks is overestimated.
In the same spirit,~\citet{le2020adversarial} propose to investigate AFLITE~\citep{sakaguchi2019adversarial}, an iterative approach to filter datasets by removing biased data points to mitigate overestimation of language models' performance. 
More recently,~\citet{alzahrani2024benchmarks} show that performance of LLMs is highly sensitive to minor changes in benchmarks with multiple-choice questions. 
Other studies demonstrate that benchmarks often include redundancy and effective LLM evaluation can be achieved with a significantly smaller sample size~\citep{polo2024, vivek2024}.

Our work is orthogonal yet complementary to previous work. In particular, we propose a new method to identify biases in a benchmark by looking at the performance of multiple recent LLMs on that benchmark.
We show that similarity in performance correlates with similarity in prompts. To the best of our knowledge, our work is the first approaching benchmark biases by analyzing and leveraging the performance of a collection of models on a set of major benchmarks; as well as  investigating the impact of inherent distributional biases in benchmarks used on LLM comparative studies.

\section{Proposed method}
\label{sec:proposed_method}

In this section, we outline the problem setup and introduce the notation and expressions that will be employed throughout the paper.  
Second, we present the approach to evaluate whether relationships between prompts (based on models' performance) are statistically non-random.  
Furthermore, we describe our method for analyzing how sensitive model comparisons are with respect to different distributional assumptions of the benchmark.
Finally, we present our proposed methodology for exploring the origins of relationships between prompt performance vectors.

\subsection{Problem setup}

Consider a benchmark containing $n$ prompts $\{p_1, \ldots, p_n\}$, and a set of $k$ LLMs $\{m_1, \ldots ,m_k\}$ being evaluated.
We define the performance matrix $\perfmatrix$ as an $n\times k$ matrix, where every cell $\perfmatrix[i,j]$ represents the performance of model $m_j$  on prompt $p_i$. 
We refer to the $i$-th row of that matrix, $\perfmatrixvector_{i}$, as a \emph{performance vector} for prompt $p_i$.
To measure how similar two prompts are with respect to model performance, 
we compute the similarity between their performance vectors $\perfsim(p_i, p_j) := s(\perfmatrixvector_{i}, \perfmatrixvector_{j})$, where $s(\cdot, \cdot)$ is a similarity function. 
Here, we consider cosine, Jaccard, and Hamming similarity.
Given a performance matrix $\perfmatrix$ and a similarity function $s$, we compute a $n\times n$ similarity matrix $\similaritymatrix_s(\perfmatrix)$, where every cell $\similaritymatrix[i, j]$ is the performance similarity for prompts $p_i, p_j$: 
$\similaritymatrix[i,j] = \perfsim(p_i, p_j)$. 


Semantic meaning from text is commonly understood through the use of embeddings. 
An embedding of a prompt is a numerical vector that contains the learned representations of semantic meaning.
Measuring semantic similarity between two prompts is achieved by measuring the distance between their embeddings.
In this paper, we use ada-2 embeddings from OpenAI\footnote{\scriptsize{\url{https://openai.com/blog/new-and-improved-embedding-model}}}. 
The ada-2 embeddings are widely used and have been proven effective in various NLP tasks. 
These embeddings have shown strong performance in assessing semantic similarity between texts~\citep{aperdannier2024systematic, kamalloo-etal-2023-evaluating-embedding, freestone2024word}. 
For a set of prompts $\{p_1, \ldots, p_n\}$, we compute a matrix of embeddings $E=\{\mathbf{e}_1, \ldots, \mathbf{e}_n\}$. 
$E$ is a $n\times s$ matrix, where $s$ is the size of the embedding vectors. 
To measure semantic similarity between pairs of prompts, we compute similarity metrics between the corresponding rows: $\semanticsim(p_i, p_j) = s(\mathbf{e}_{i}, \mathbf{e}_{j})$.

\subsection{Determining if performance vectors are correlated}
\label{subsec:method_prompt_correlation}

Given a benchmark, we assess whether the observed similarity among performance vectors is significant.
If the observed similarity is significantly high, this implies the existence of specific connections between prompts. 
These connections lead to similar model behavior when responding to these prompts.

To test this hypothesis, we perform permutation tests.
We generate permutations of the performance matrix $\perfmatrix$ by randomly shuffling the cells of each column. 
In this way, we permute the values of the model responses across prompts, while holding constant the overall performance of each model (i.e., the column averages of $\perfmatrix$).
We then compute a similarity matrix $\similaritymatrix_s(\perfmatrix)$ for the observed performance matrix $\perfmatrix$, as well as for each permutation $\perfmatrix'$ of the performance matrix: $[\similaritymatrix_s(\perfmatrix'_1), \similaritymatrix_s(\perfmatrix'_2), \ldots]$.
We compare the distribution of values from $\similaritymatrix_s(\perfmatrix)$ with the distribution of values from the permuted tables $[\similaritymatrix_s(\perfmatrix'_1), \similaritymatrix_s(\perfmatrix'_2), \ldots]$. 
We conduct a permutation test to compare the average, 75th, and 95th percentiles of these distributions.
The p-value of the permutation test is calculated as the proportion of permuted tables for which the statistic is greater than the one obtained with the observed table.
Additionally, we use the Kolmogorov-Smirnov (KS) test to compare the entire distribution of values between observed and permuted similarity matrices.

To further support our findings, we cluster the observed and permuted performance vectors.
If there are non-random correlations between performance vectors, we would expect the clustering of the observed vectors to have higher clustering quality metrics, such as silhouette score.

\subsection{Effect of non-uniform weights in aggregate performance metrics}
\label{subsec:random_weights}

So far, we have focused on aggregate performance measures that treat prompts as if they are independent and identically distributed (i.i.d.)\ samples from some real-world distribution of interest---i.e., each prompt is given equal weight in calculating aggregate performance metrics. 
In this section, we examine the implications of relaxing this assumption for ranking models based on their performance. 
Generally, there is no universally correct distribution of interest---it depends on each user's application.
Here, we look into three different ways of capturing distributional assumptions (i.e., of defining weights) for a given benchmark. 

\paragraph{Cluster-based:}
We leverage the clustering of performance vectors described above. 
We consider the following variants for evaluating performance:
\begin{asparaenum}
\item
Only include prompts that are cluster representatives (i.e., the medoids of the clusters). 
This effectively decreases the size of the benchmark. 
\item Include all prompts, but weigh them based on their distance from their cluster representative. 
We employ two types of weights: 
\begin{asparaenum}[(i)]
\item \emph{Distance-based:} The further away a prompt is from the cluster representative, the larger its weight. 
This setting gives more emphasis on diversity of the benchmark. 
More formally, let $p_i$ be a prompt in cluster $C_j$, $p^r_j$ be the representative prompt of cluster $C_j$, and $d(\cdot,\cdot)$ the distance function between two prompts. 
The weight $w$ for $p_i$ is: 
\[
w(p_i) = \frac{d(p_i, p^r_j)}{ \sum_{p_k\in C_j}\left(d(p_k, p^r_j)\right)} \frac{|C_j|}{\sum_i |C_i|}
\]
The first factor is the within-cluster weight of the prompt (normalized within cluster). 
The second factor weighs all prompts of a given cluster proportionally to the cluster's size. 
\item \emph{Inverse-distance weights:} The closer a prompt is to the cluster representative, the larger its weight. 
This setting effectively smooths out the hard clustering we produced: all data points contribute to the performance, not just the cluster representatives. 
The weight $w$ for $p_i$ is computed as: 
\[
w(p_i) = \frac{d^{-1}(p_i, p^r_j)}{ \sum_{p_k\in C_j}\left(d^{-1}(p_k, p^r_j)\right)} \frac{|C_j|}{\sum_i |C_i|}
\]
\end{asparaenum}
\end{asparaenum}

\paragraph{Increasing benchmark size}
We start with a random prompt and iteratively add new prompts into the benchmark. 
To select the next prompt to add, we use two methods: 
\begin{inparaenum}[(i)]
\item \emph{most informative:} select the prompt with the largest cosine distance (lowest cosine similarity) from the previously selected ones in order to obtain an informative test set with a reduced semantic similarity between prompts,
\item \emph{random:} select a random prompt. 
\end{inparaenum}

\paragraph{Random distributions of weights}
We weigh each prompt and compute weighted performance, with weights drawn uniformly at random.
To achieve that, we sample uniformly at random from the unit simplex using the sampling technique described in~\citet{smith2004sampling}.
This approach aims to provide a characterization over all possible weight configurations. 

\subsection{Comparing performance vectors with semantic embeddings of prompts} 
\label{subsec:semantic_similarity_method}

Having established that model performance is similar across prompts, we next investigate where this similarity stems from. 
Our hypothesis is that for a pair of prompts, similar model performance can occur if the prompts are semantically similar. 

We use linear regression to determine if there exists a significant relationship between semantic similarity and model performance similarity: 
$$\perfsim(p_i, p_j) = \semanticsim(p_i, p_j)\beta + \epsilon$$
where  $\beta$ is the coefficient of how much semantic similarity contributes to the model and $\epsilon$ is error.

Using all prompt pairs raises concerns about the data being i.i.d., given that each observation is a pairwise comparison and each member of a pair appears in many observations. 
To avoid that, we estimate one model for each prompt, including all the pairwise observations of which that prompt is a part. 
We collect p-values for the coefficients across all models and perform multiple hypotheses adjustment to generate False Discovery Rate (FDR) values. 
We repeat the same approach for 1000 permutations as described in Section~\ref{subsec:method_prompt_correlation} for both pairwise performance and semantic similarity vectors. 
Finally, we compare the distribution of coefficients and FDRs between original data and permutations using the KS test.

\section{Experimental setup}
\label{sec:experimental_setup}

In this section, we describe the setting of our experiments. 
Specifically, we provide details on the benchmarks and evaluation metrics we use, the LLMs we consider, and how we evaluate performance of the LLMs on the benchmarks.

\subsection{Benchmarks}

We investigate four major benchmarks that are designed for different tasks.

\paragraph{ANLI} The Adversarial Natural Language Inference (ANLI) dataset\footnote{\scriptsize{\url{https://huggingface.co/datasets/anli}}} is a large-scale dataset for natural language inference (NLI)~\citep{nie2020adversarial}.
It is collected via an iterative, adversarial human-and-model-in-the-loop procedure, making it more difficult than its predecessors.
The dataset used here comprises approximately 100K samples for the training set, 1,200 for the development set, and 1,200 for the test set.
Each sample contains a context, a hypothesis, and a label. 
The goal is to determine the logical relationship between the context and the hypothesis. 
The label is the assigned category indicating that relationship. 
In the context of NLI, the labels typically include ``entailment'', ``contradiction'', or ``neutral''. 
Finally, ANLI makes available a reason (provided by the human-in-the-loop), explaining why a sample was misclassified.

\paragraph{HellaSwag}

This is a commonsense natural language inference dataset \citep{zellers2019hellaswag}, tasking machines with identifying the most probable followup for an event description. 
Comprising 70,000 instances, each scenario presents four potential outcomes, with only one being accurate. 
Engineered to be challenging for cutting-edge models, the dataset employs Adversarial Filtering to incorporate machine-generated incorrect responses, frequently misclassified by pretrained models. 
Covering diverse domains, HellaSwag demands a fusion of world knowledge and logical reasoning for successful interpretation.

\paragraph{CommonsenseQA}
This is a multiple-choice question-answering dataset that requires different types of commonsense knowledge to predict the correct answers \citep{talmor2019commonsenseqa}. 
It contains 12,102 questions with one correct answer and four distractor answers. 
The questions are crowdsourced and cover a wide range of topics such as open-domain-qa, real-life situations, elementary science, social skills. 

\paragraph{CNN/Daily Mail}
The CNN/Daily Mail dataset is a widely used benchmark for text summarization~\citep{nallapati-etal-2016-abstractive-new}. The dataset comprises news stories from CNN and Daily Mail websites. 
In total, the corpus contains 286,817 training, 13,368 validation, and 11,487 test pairs.

\subsection{Evaluation measures}
\label{subsec:evaluation_measures}
For ANLI, HellaSwag, and CommonsenseQA, the performance matrix contains binary values (correct / incorrect answer). Hence, we use average accuracy to evaluate the performance of each model, as commonly done with these benchmarks~\citep{nie2020adversarial,wei2022finetuned, zellers2019hellaswag,talmor2019commonsenseqa}.
For CNN/Daily Mail, following previous work~\citep{see2017get}, we measure model performance using the ROUGE score. 

\subsection{Considered LLMs}
\label{subsec:llms}

In order to have a diverse collection of LLMs, we include models from several developers, such as OpenAI and Meta.
These include GPT LLMs~\citep{brown2020language,openai2023gpt}, Llama LLMs~\citep{touvron2023llama}, and other popular LLMs, such as Falcon-180b~\citep{falcon}, Koala 13B~\citep{geng2023koala}, Alpaca 7B~\citep{wang2023far}.
Table~\ref{tab:llms_datasets} shows the various models used for each benchmark\footnote{Due to constraints in LLMs' availability, we use different LLMs for each benchmark. This does not impact our work, as each benchmark analysis is standalone and independent of the remaining benchmarks.}.

\subsection{Performance evaluation}
\label{subsec:performance_evaluation}
 
For ANLI, we evaluate each model on the test dataset, which contains 1200 prompts. 
For each sample, we use 7 few-shot samples extracted from the ANLI dev set. 
For the remaining benchmarks, we randomly sample 10\% of each benchmark for test and use the rest for few-shot selection. 
This results in 1005, 1221, and 1150 test samples for HellaSwag, CommonsenseQA, and CNN/Daily Mail respectively.
For HellaSwag, we use 10 few-shot examples, while for CommonsenseQA and CNN/Daily Mail we use 5 few-shots.


{\setlength\tabcolsep{0.5pt}
\begin{threeparttable}[tb]
\footnotesize
    \centering
    \begin{tabular}{|c|l|c|c|c|c|}
        \hline
      Type   &Model  & ANLI & HS & CSQA & CNN/DM \\
        \hline
      \multirow{16}{*}{\rotatebox[origin=c]{90}{GPT}}   & ChatGPT-Turbo-Base-0516 & \checkmark & \checkmark & & \\
        & ChatGPT-Turbo-0301& \checkmark & \checkmark & & \\
        & ChatGPT-Turbo-0613 & & & \checkmark & \\
        & ChatGPT-202301 & & & & \checkmark \\
        & DaVinci (GPT-3) & & & \checkmark & \\
        & Text-Davinci-002 & & & & \checkmark \\
       &  Text-Davinci-003 & & & & \checkmark \\
        & GPT-4-0314 & & & \checkmark & \\
        & GPT-4-0314 (Chat) & \checkmark & \checkmark & &  \checkmark\\
        & GPT-4-0613 (Chat) & & & \checkmark & \\
        & GPT-4-Turbo-1106 (Chat) & \checkmark &\checkmark & \checkmark & \\
       &  GPT-4-Turbo-1106 & & & \checkmark & \\
       &  Text-Alpha-002-Current & & \checkmark & & \checkmark \\
       &  DV3-FP8 & & & & \checkmark \\
        & Babbage-0721 & & & & \checkmark \\ 
        & ChatGPT-202301 & & & & \checkmark \\
        \hline
        \hline
      \multirow{5}{*}{\rotatebox[origin=c]{90}{LLAMA}}   & Llama-13B & & & \checkmark & \\
         &Llama-2-13B & \checkmark & \checkmark & & \\
       &  Llama-30B  & & \checkmark & \checkmark & \\
        & Llama-65B & & & \checkmark & \\
        & Llama-2-70B & \checkmark & \checkmark & \checkmark & \\
        \hline \hline
     \multirow{7}{*}{\rotatebox[origin=c]{90}{Other}}      & Persimmon 8B\tnote{1} & \checkmark & \checkmark & \checkmark & \\
       &  Vicuna 13B\tnote{2} & \checkmark & \checkmark & & \\
        & Claude-2\tnote{3} & \checkmark & \checkmark & \checkmark & \\
       &  Falcon-180b & \checkmark & & \checkmark & \\
         & Koala 13B & \checkmark & \checkmark & & \\
         &Mistral7b\tnote{4} & \checkmark & & \checkmark & \\
       &  Alpaca 7B &  & \checkmark & & \\
         \hline
         \hline
        & Total  & 12 & 13 & 14 & 8 \\
        \hline
    \end{tabular}
    \begin{tablenotes}\scriptsize
    \item[1] \url{https://www.adept.ai/blog/persimmon-8b}
    \item[2] \url{https://lmsys.org/blog/2023-03-30-vicuna/}
    \item[3] \url{https://www.anthropic.com/index/claude-2}
    \item[4] \url{https://mistral.ai/news/announcing-mistral-7b/}
    \end{tablenotes}
    \caption{Summary of LLMs used for ANLI, HellaSwag (HS), CommonsenseQA (CSQA), and CNN/Daily Mail (CNN/DM).
    Check marks denote which LLMs were used for the specific benchmark.}
    
    \label{tab:llms_datasets}
\end{threeparttable}
}

\section{Results}

In this section, we present the results of the experiments described in Section~\ref{sec:proposed_method} on the benchmarks.

\subsection{Performance vectors are correlated}
\label{subsec:model_response_correlation}

To determine if prompt performance vectors are correlated, we perform the permutation tests described in Section~\ref{subsec:method_prompt_correlation}, using different correlation measures.
The obtained p-values for ANLI, HellaSwag, and CommonsenseQA are depicted in Table~\ref{tab:model_comparison_pvalue_all}.
On ANLI and CommonsenseQA, the permutation tests show strong evidence that the correlations between the prompt performance vectors are significant. 
For HellaSwag, our findings reveal consistently low p-values across all correlation measures when using the 75th percentile, as well as a low p-value when averaging Jaccard similarities. 
For the three benchmarks above, the KS test is significant across all correlation measures. 

\begin{table}
\small
    \centering
    \begin{tabular}{l|llll}
        \hline
        && Hamming  & Cosine & Jaccard \\
        \hline
        \multirow{4}{*}{\rotatebox[origin=c]{90}{ANLI}} &Average  &0.60  &0.59 & \textbf{0.0009}\\
        &75th percentile  &0.66  & \textbf{0.0009} &0.67 \\
        &95th percentile &\textbf{0.0009}  & \textbf{0.0009} & \textbf{0.0009}\\
        & KS test &\textbf{2e-5}  & \textbf{2e-5} & \textbf{2e-5}\\
        \hline
        \multirow{4}{*}{\rotatebox[origin=c]{90}{HS}} &Average  &0.52&0.57&\textbf{0.0009}\\
        &75th percentile  &\textbf{0.0009}&\textbf{0.0009}&\textbf{0.0009}\\
        &95th percentile  &0.88&0.85&0.87\\
         &KS test &\textbf{2e-5}  & \textbf{2e-5} & \textbf{2e-5}\\
         \hline
        \multirow{4}{*}{\rotatebox[origin=c]{90}{CSQA}} &Average  &0.53&0.52&\textbf{0.0009}\\
        &75th percentile  &\textbf{0.0009}&\textbf{0.0009}&\textbf{0.0029}\\
        &95th percentile  &\textbf{0.0009}&\textbf{0.0009}&\textbf{0.0009}\\
        & KS test &\textbf{2e-5}  & \textbf{2e-5} & \textbf{2e-5}\\
        \hline
    \end{tabular}
     \caption{p-values obtained with permutation tests and the KS test using different correlation measures and aggregation functions for ANLI, HellaSwag (HS), and CommonsenseQA (CSQA).}
    \label{tab:model_comparison_pvalue_all}
\end{table}

For CNN/Daily Mail
the performance matrix contains ROUGE scores, which are continuous values. 
Thus, we use cosine similarity to compare the average correlations obtained from the original and permuted performance matrices.
The results show that the correlations among original performance vectors are significantly greater. 

To further support this finding, we cluster the model responses using spherical $k$-means~\citep{dhillon2001concept}. 
We choose the optimal number of clusters to maximize the average silhouette score, computed using cosine distance. 
Table~\ref{tab:silhouette_scores} contains the average silhouette scores of clustering the performance vectors and a random permutation of them. 
For all benchmarks, the performance vectors produce higher silhouette scores compared to the permuted performance vectors.
This provides additional evidence to support the outcome of the hypothesis tests presented above: the performance vectors are similar.


\begin{table}
\small
    \centering
    \begin{tabular}{lll}
          \hline
         Benchmark & observed & permuted\\
         \hline
         ANLI & \textbf{0.52} & 0.21\\
         HellaSwag & \textbf{0.54} & 0.24 \\
         CommonsenseQA & \textbf{0.61} & 0.29\\
         CNN/Daily Mail & \textbf{0.25} & 0.21\\
         \hline
    \end{tabular}
    \caption{Average silhouette score of clustering observed performance vectors and a random permutation of performance vectors for the various benchmarks.}
    \label{tab:silhouette_scores}
\end{table}

\subsection{Impact of prompt weights on performance and relative ranking of models}

In this section, we present the results of different weighting schemes for the prompts of a benchmark, as described in Section~\ref{subsec:random_weights}.

\subsubsection{Cluster-based evaluation}

\begin{figure*}
    \centering
    \subcaptionbox{\label{subfig:anli_ranking_medoids} ANLI}
    {\includegraphics[width=.48\textwidth]{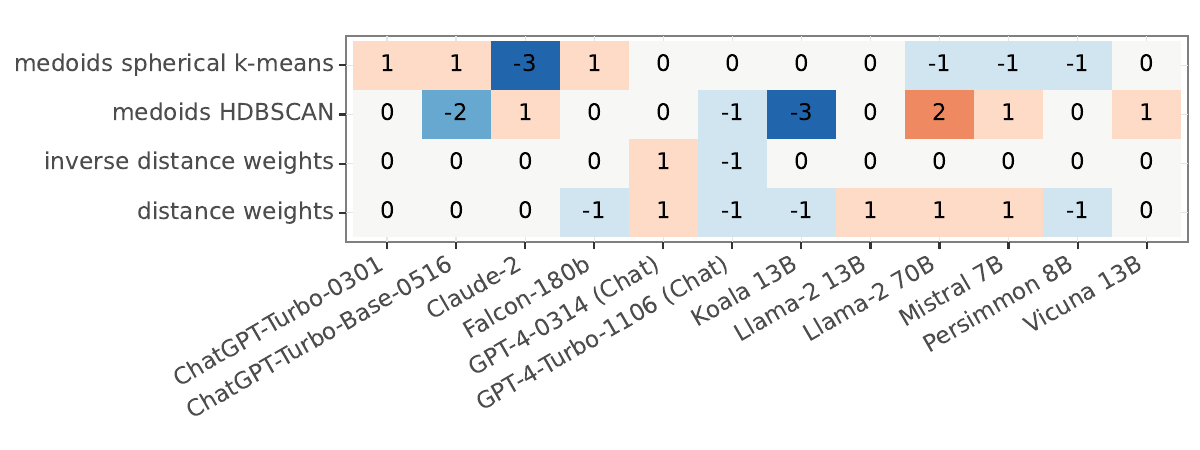}}
%
    \subcaptionbox{\label{subfig:hellaswag_ranking_medoids} HellaSwag}
    {\includegraphics[width=.48\textwidth]{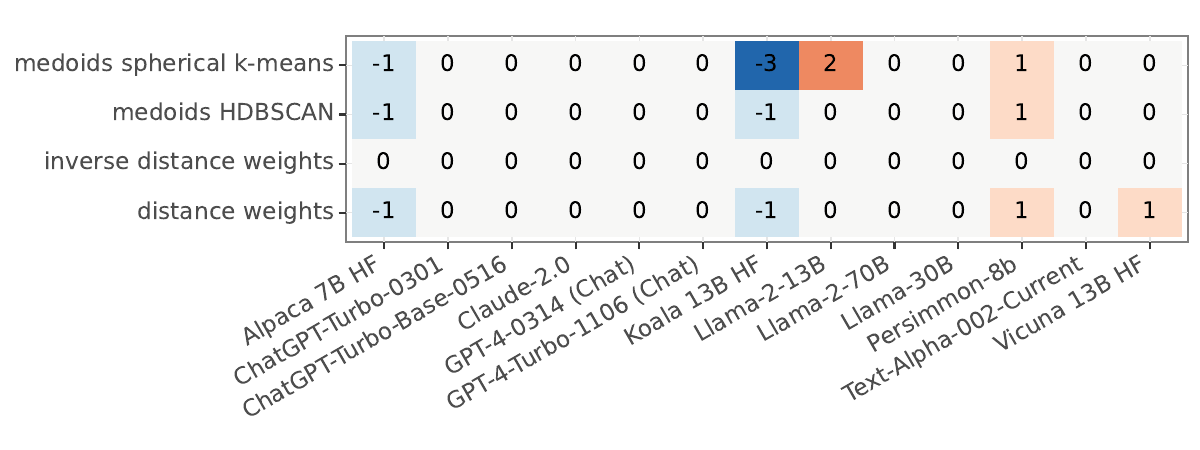}}
%
    \subcaptionbox{\label{subfig:CommonsenseQA_ranking_medoids} CommonsenseQA}
    {\includegraphics[width=.48\textwidth]{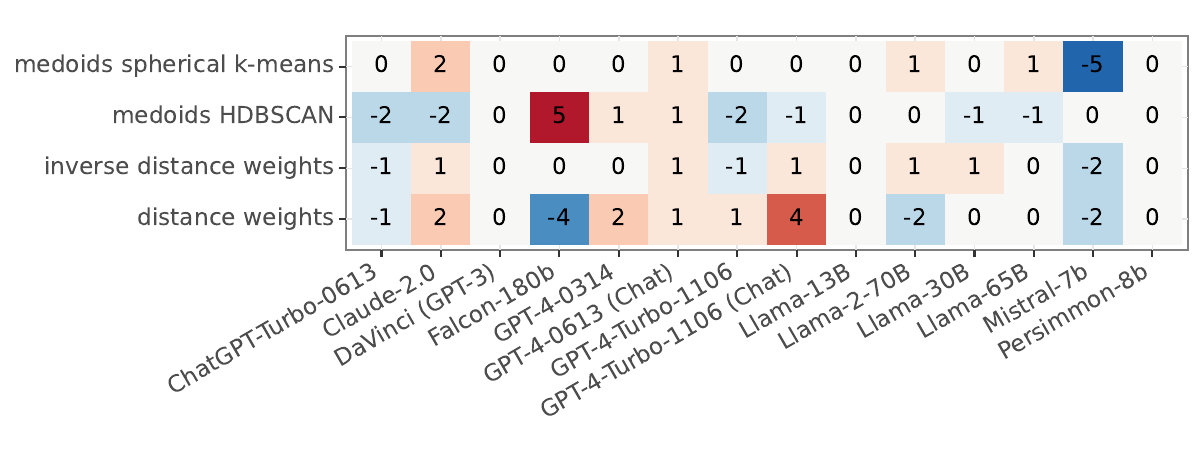}}
%
    \subcaptionbox{\label{subfig:cnn_ranking_medoids} CNN/Daily Mail}
    {\includegraphics[width=.48\textwidth]{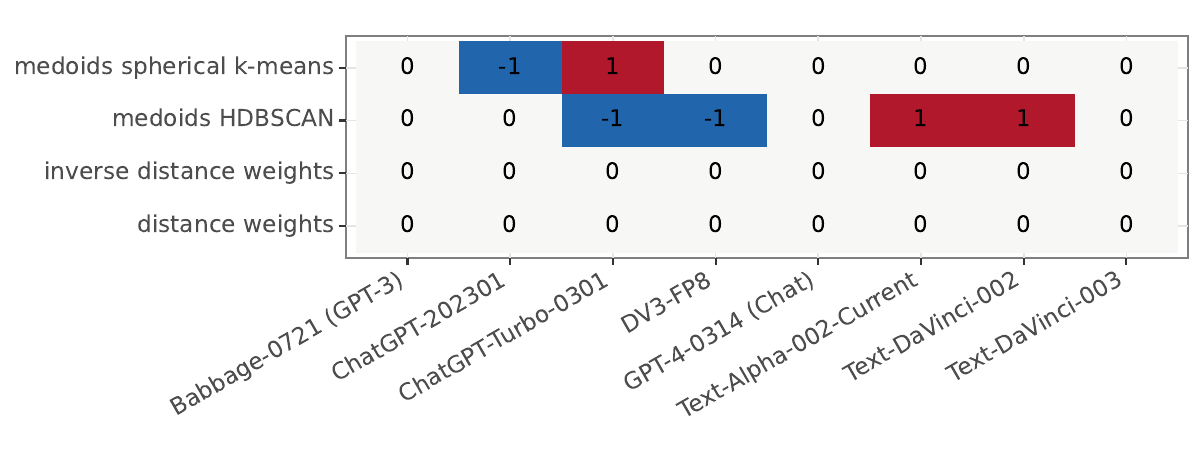}}
\caption{Visualization of ranking changes (compared to original benchmark) for various benchmark modifications. 
Rows show different weighting methods, columns show the models. 
Each cell contains the ranking change (original ranking minus new ranking)
 of the column-model for the row-method. 
 We observe rank changes as great as 5.}
\label{fig:rankings_medoids}
\end{figure*}

First, we cluster the performance vectors of each benchmark as described earlier. 
Then, we compute the average accuracy of models for each benchmark, using only the cluster representatives of that benchmark.
We also compute weighted performance using distance-based and inverse-distance-based weights. 
Figure~\ref{fig:rankings_medoids} illustrates how these weighting schemes affect the relative ranking of models for each benchmark.
The rows correspond to different weighting schemes, while the columns correspond to the different models and are ordered by increasing original performance (i.e., decreasing rank).
Every cell contains the ranking change (compared to the original benchmark) of the model of that column for the method of that row. 
If there were no ranking changes, all values would be 0.
However, we observe that there are multiple ranking changes as great as 5 (model is ranked 5 positions above the original benchmark).

\subsubsection{Increasing size of benchmark}
\label{subsec:increasing_size_benchmark}

\begin{figure}
    \centering
    \subcaptionbox{\label{subfig:performance_mmr} ANLI}
    {\includegraphics[width=.48\textwidth]{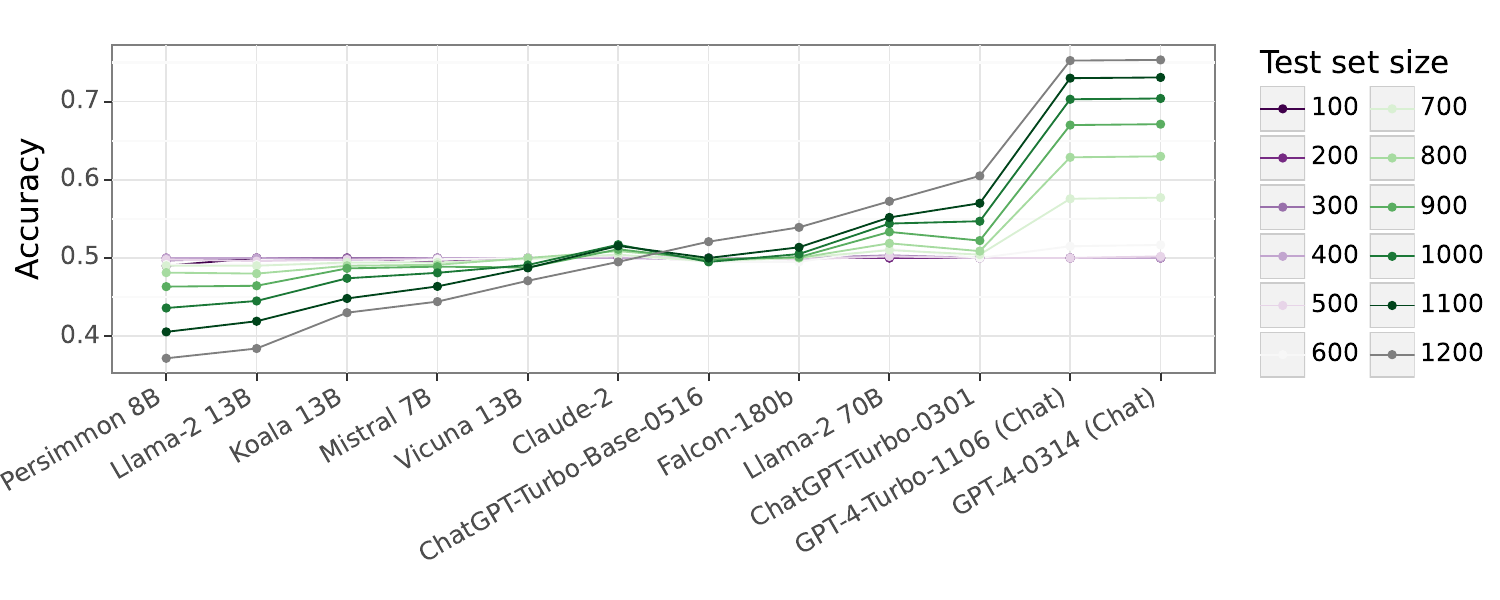}}
    \subcaptionbox{\label{subfig:cnn_performance_mmr} CNN/Daily Mail}
    {\includegraphics[width=.48\textwidth]{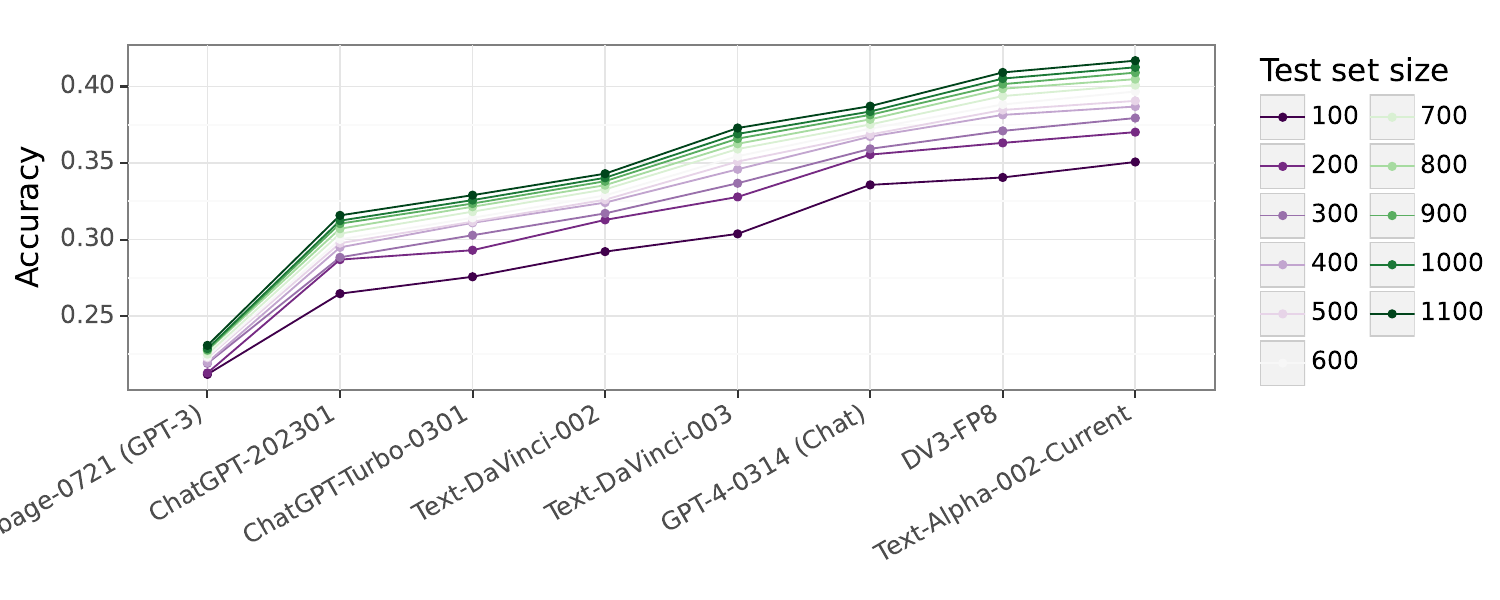}}
\caption{Average performance as benchmark size increases. Prompts are added to maximize average cosine distance. Maximum benchmark size corresponds to performance on the original benchmark.}
\label{fig:performance_mmr}
\end{figure}

Next, we study how performance is affected by the size and diversity of the benchmark. 
We start with a random prompt and iteratively add new prompts to the benchmark, either by adding the most informative prompt (i.e., the one with the maximum average distance from the current benchmark), or a random one.
Figure~\ref{fig:performance_mmr} shows the average performance for each model as the benchmark size increases 
(maximum benchmark size corresponds to the original benchmark).
Looking at the most informative method for ANLI (Figure~\ref{subfig:performance_mmr}), the first 400 prompts result in random performance (0.5) for all models. 
This suggests that the initial prompts chosen with this method are the most ``difficult'', in that the models are exhibiting performance close to random (accuracy 50\%). 
Similar results are observed for HellaSwag and CommonsenseQA (see Appendix~\ref{sec_app:benchmark_size_increase}, Figure~\ref{fig:performance_mmr_appendix}), but not for CNN/Daily Mail (Figure~\ref{subfig:cnn_performance_mmr}), where the performance on the reduced benchmark follows a similar pattern as the performance on the original benchmark.
The random method tracks the original performance for all benchmarks (see Appendix~\ref{sec_app:benchmark_size_increase}, Figure~\ref{fig:performance_random_appendix}).

\subsubsection{Random distributions of weights}
\label{subsubsec:random_weights}

We explore the distribution of all weighting schemes and the effect they have on the weighted accuracy and relative ranking of the models.
As described in Section~\ref{subsec:random_weights}, we sample 100,000 random weight configurations. 
For each model, we compute the weighted performance based on these weights.

For ANLI, HellaSwag, and CommonsenseQA the performance of a model can change up to 10\%. 
For CNN/Daily Mail, the range is smaller, up to 3\%. 
Detailed results are included in Appendix~\ref{sec_app:random_weights_plots}.
We note that the range is similar for all models within a benchmark, indicating that it is a property related to the benchmark and not the specific models. 


    
\begin{figure}
    \centering
    \subcaptionbox{\label{subfig:random_weights_confusion_matrix} ANLI}
    {\includegraphics[width=.98\columnwidth]{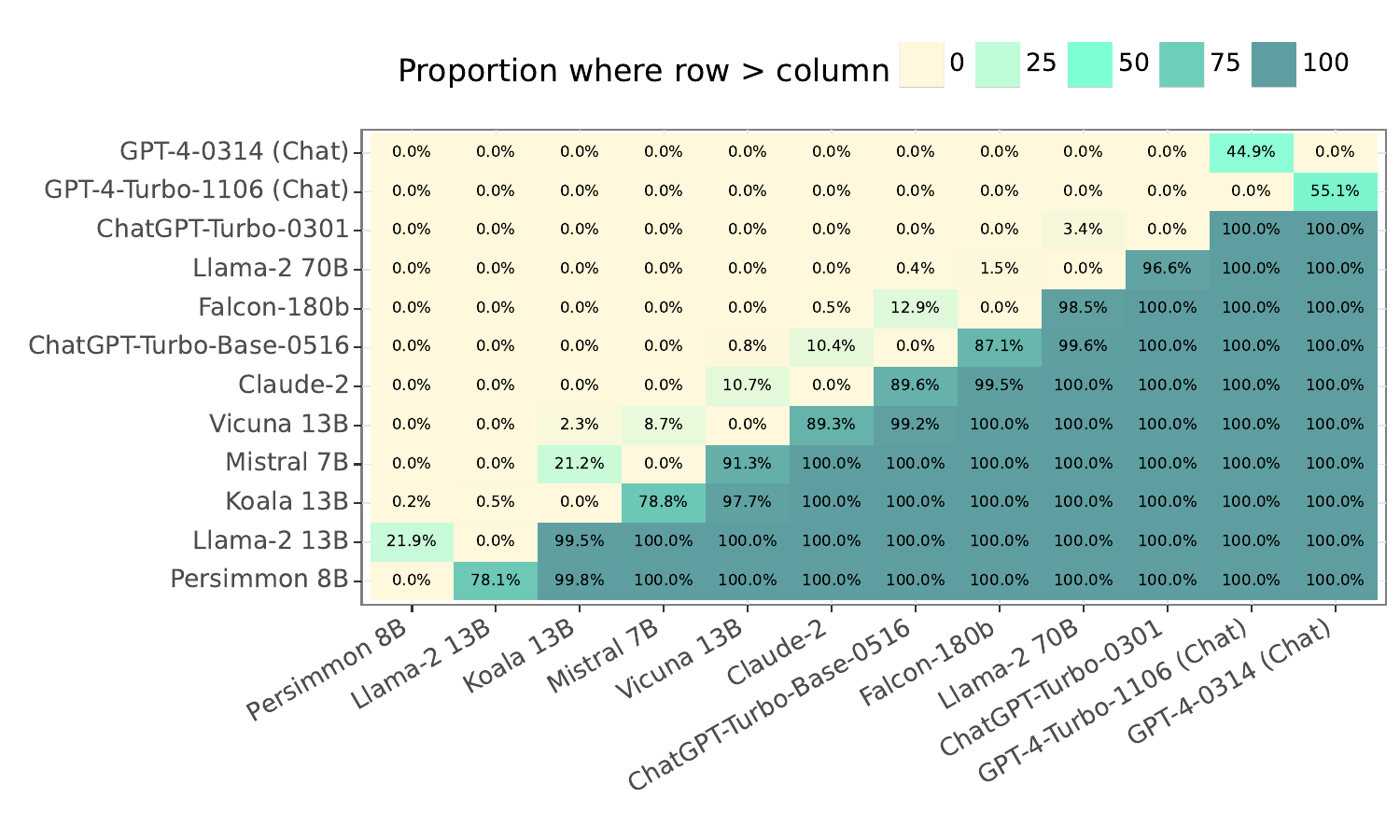}}

    \subcaptionbox{\label{subfig:cnn_random_weights_confusion_matrix} CNN/Daily Mail}
    {\includegraphics[width=.98\columnwidth]{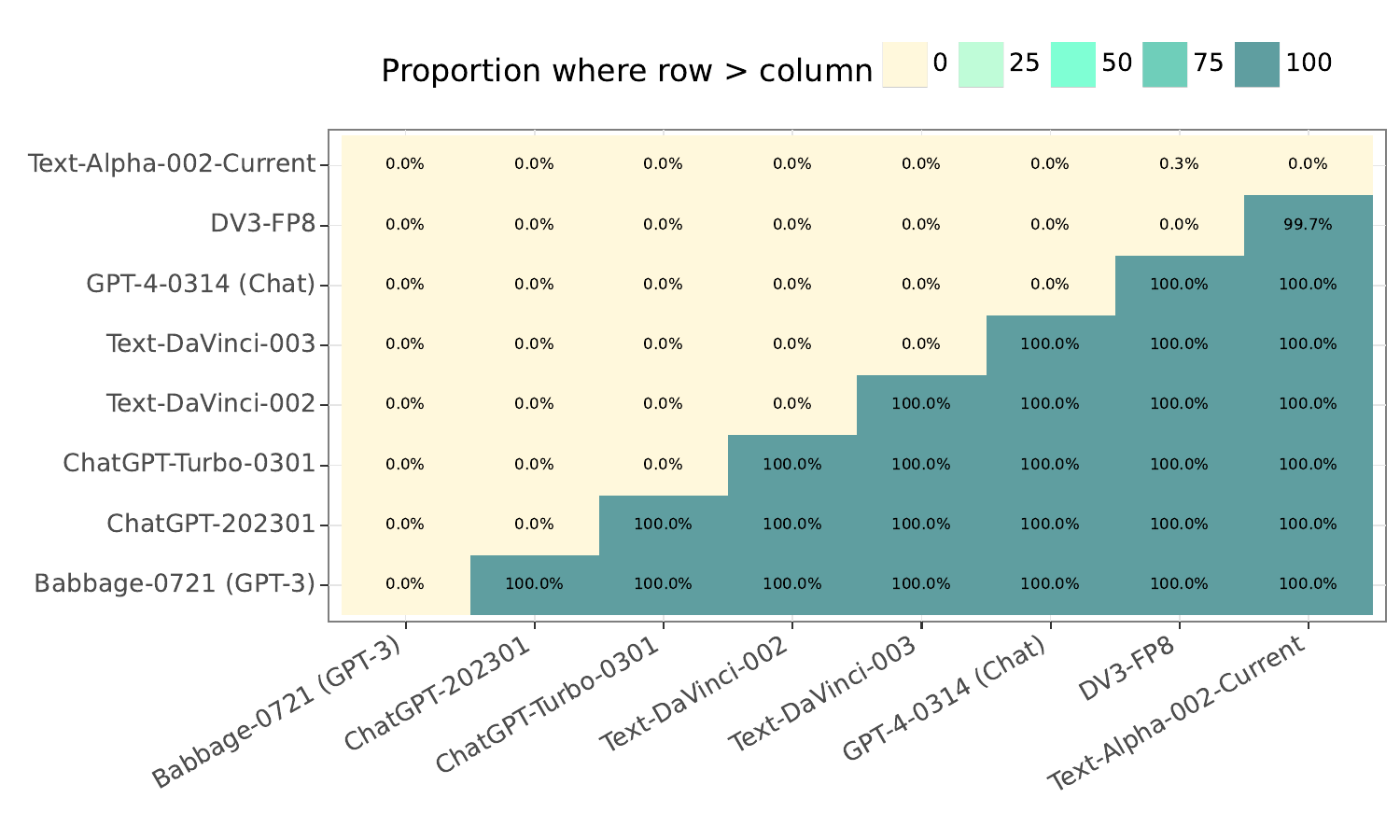}}
\caption{Pairwise comparison of weighted performance. Each cell is the percentage of times the model of the row outperforms the model of the column.}
\label{fig:random_weights_heatmap}
\end{figure}

To further demonstrate changes in relative ranking of models, we take a closer look at the pairwise ranking differences.
Figure~\ref{fig:random_weights_heatmap} depicts a pairwise comparison of weighted performance for each benchmark.
Every cell shows how often the model in the row outperforms the model of the column. 
For ANLI, approximately for half of the weight configurations the ranking of the top two models
is reversed!
However, for the CNN/Daily Mail data, there are effectively no reversals (less than 0.01\%).

\subsection{Relationship between model performance and semantic similarity of prompts}

Having established that model performance is correlated across prompts, we investigate what can explain these correlations.
Our hypothesis is that it is driven by semantic similarity. 
We use the method described in Section~\ref{subsec:semantic_similarity_method} to assess if there
is a significant relationship between semantic similarity and model performance similarity. 

\begin{figure}
    \centering
    \subcaptionbox{\label{subfig:regression_anli_test} ANLI (reason)}
    {\includegraphics[width=.48\columnwidth]{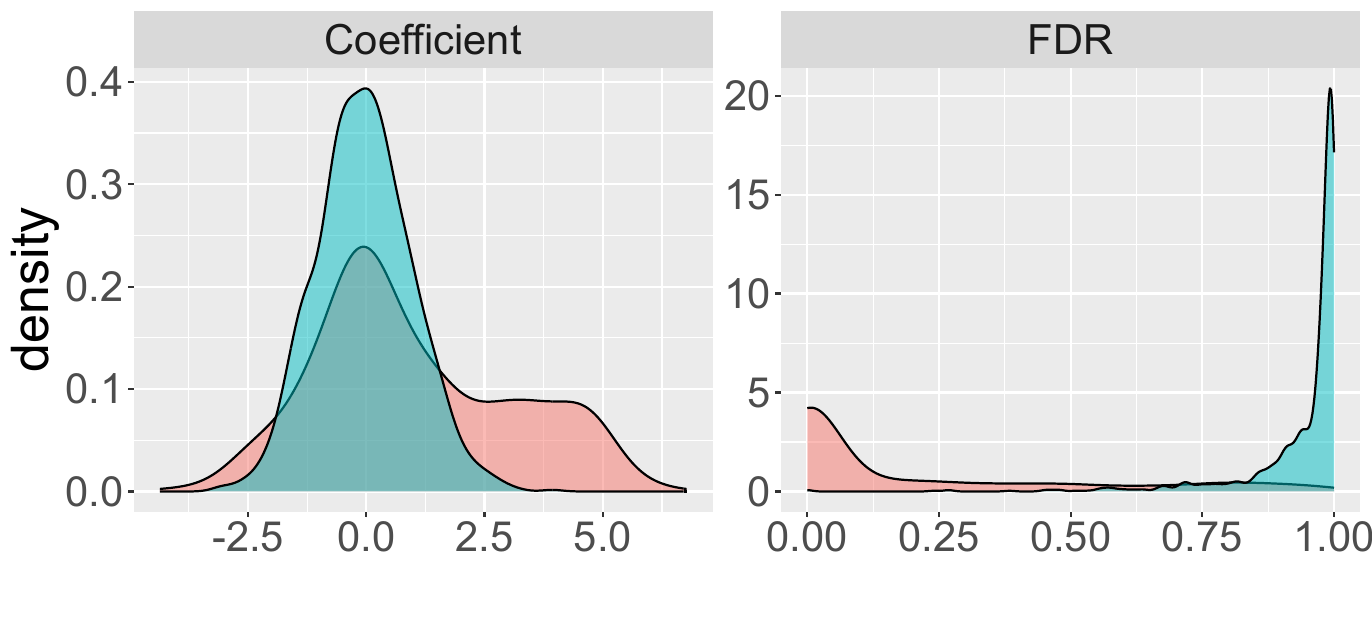}}
    \subcaptionbox{\label{subfig:regression_hellaswag_test} HellaSwag}
    {\includegraphics[width=.48\columnwidth]{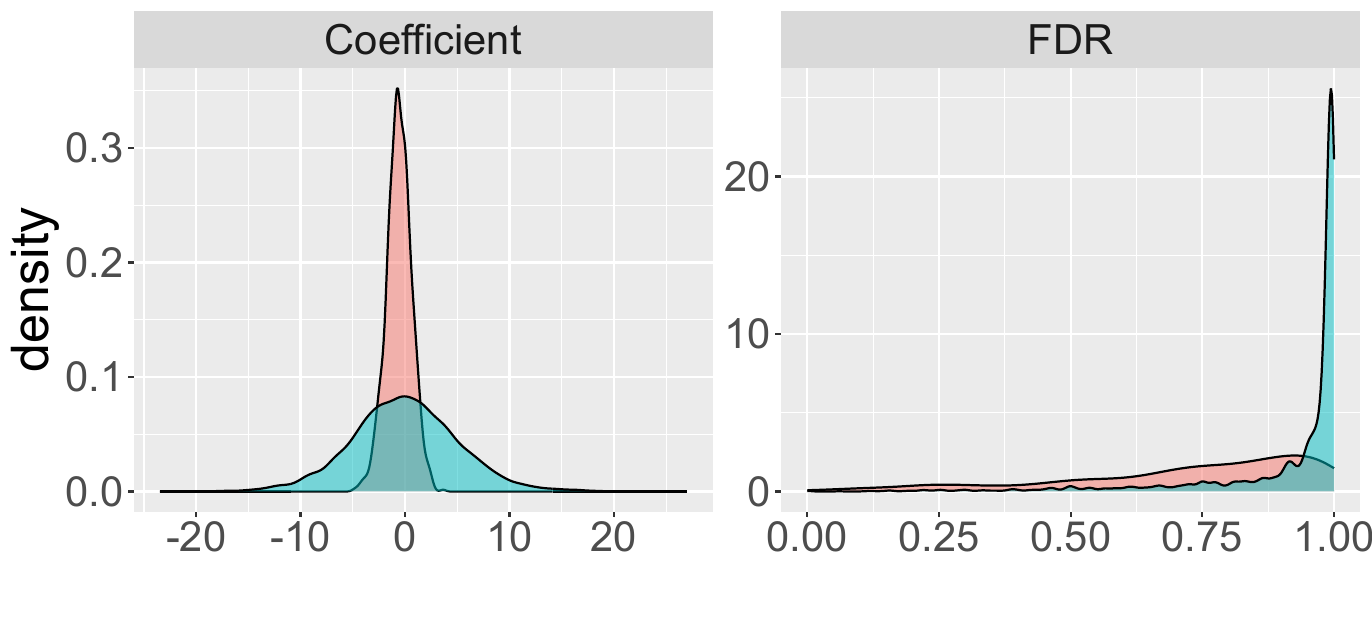}}
    \subcaptionbox{\label{subfig:regression_commonsense_test} CommonsenseQA}
    {\includegraphics[width=.48\columnwidth]{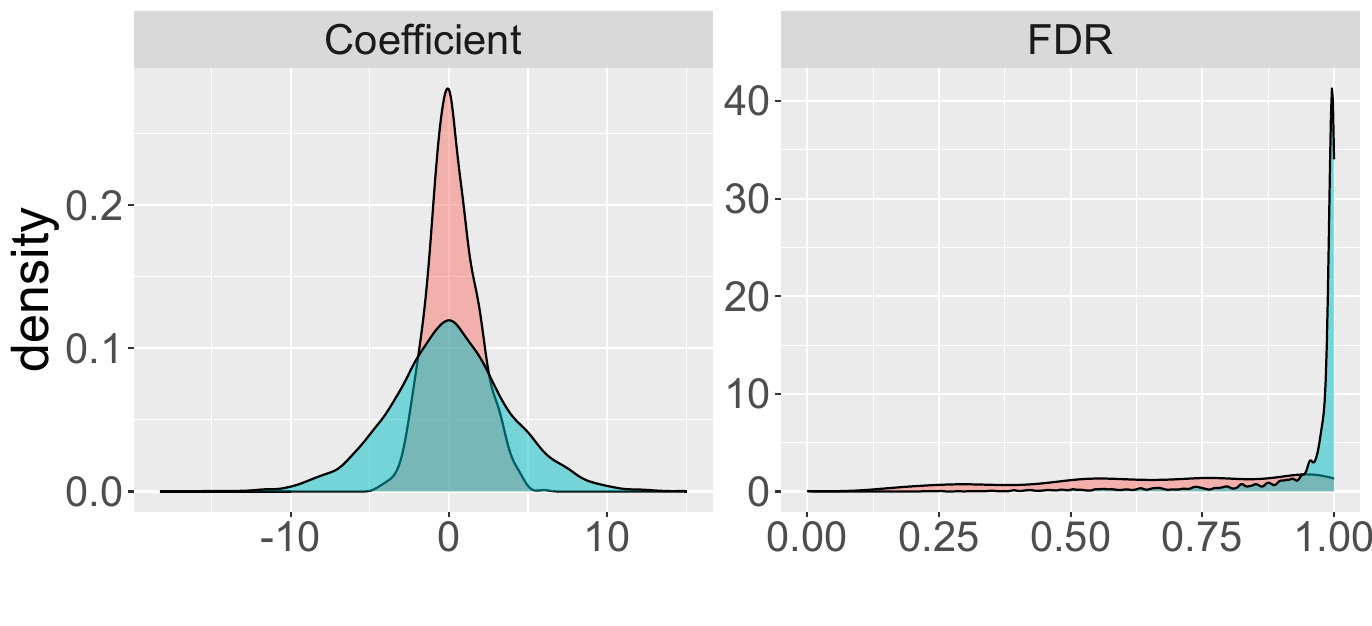}}
    \subcaptionbox{\label{subfig:regression_dailycnn_test} CNN/Daily Mail}
    {\includegraphics[width=.48\columnwidth]{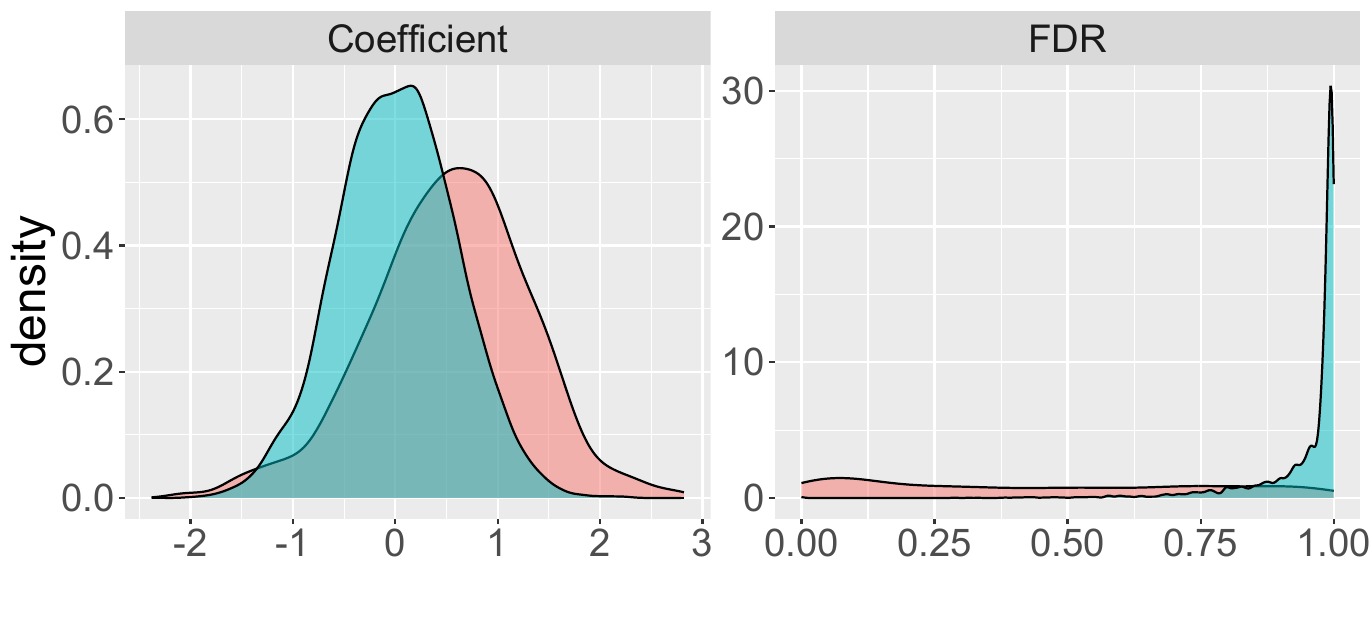}}
\caption{Distribution of semantic similarity coefficients and FDRs for all benchmarks. Red is original data, blue is permutations. KS tests for all distributions shown have p-values < 2e-5.}
\label{fig:regression_test}
\end{figure}

Our findings show that only CNN/Daily Mail presents a significant relationship between prompt semantic similarity and prompt performance similarity (see Figure~\ref{subfig:regression_dailycnn_test}).
This benchmark is a text summarization task, where the success of the ROUGE metric highly depends on the ability to extract relevant entities from text. 
For example, we find that prompts referring to the economy or global warming have high correlation in model performance (see Appendix~\ref{sec_app:similar_prompts}, Table~\ref{tab:dailycnn_regression_reason_example}).

ANLI also makes available a reason component: what human agents state as the explanation for why the LLM gave a wrong answer.
We find a significant relationship between semantic similarity using the reason component and prompt performance similarity
(as seen in Figure~\ref{subfig:regression_anli_test}).
The input prompt---consisting of the context, hypothesis and label components---shows no relationship, which is most likely because the creators of ANLI put great effort into ensuring diversity in the benchmark~\citep{nie2020adversarial}. 
This is also evident in Figure~\ref{fig:performance_mmr}.
The significance of the reason component indicates that the model performance vectors correlate because of \emph{how} the model generates a response. 
We observe prompts where the reasons for similar model performance indicate that the model cannot do math, e.g., ``The system may have
missed this as it did not add up the losses from both sets'' and ``the model might not know math'' (see Appendix~\ref{sec_app:similar_prompts}, Table~\ref{tab:anli_regression_reason_example}).

Hellaswag and CommonsenseQA use a multiple-choice format. 
The lack of strong evidence supporting the correlation in these benchmarks (see Figures~\ref{subfig:regression_hellaswag_test} and~\ref{subfig:regression_commonsense_test}) is likely due to the embeddings picking up similarities between the different choices, rather than the logic the LLMs employ to arrive at their conclusion.
This is consistent with our findings for ANLI, where a significant relationship does not stem from inputs to the model, but from the LLMs’ failure points.

Our findings indicate there is a larger question about why the model performance vectors are correlated, and investigating this is central to understanding model performance.
Semantic similarity can be a factor, but it depends on the task the benchmark is designed for. 
Based on our results for ANLI, it appears that the reasoning required for the task (i.e., reasoning types that cause models to fail), can be even more important than semantic similarity.

\section{Conclusion and future work}

LLMs are commonly evaluated on benchmarks that may include multiple prompts testing similar skills. 
In this work, we demonstrate this bias on major benchmarks, by showing that model performance across different prompts is significantly correlated. 
Furthermore, we demonstrate that LLM comparative studies can be significantly altered when using non-uniform weights for prompts during evaluation. 
The suggested approach can serve as a consistency check in comparative studies of LLMs, ensuring that the results take into consideration benchmark biases. 
Finally, we show that similar model performance across prompts can be explained by semantic similarity, but is most likely derived from common failure points of the LLM.



Our findings could influence a larger diagnostics tool for evaluating the robustness of model quality comparisons with respect to distributional assumptions of benchmarks.
Future work also includes identifying additional factors that may explain these biases. 
This information can give rise to solutions for improving benchmarks robustness. 
These findings could help researchers generating novel benchmarks to identify and eliminate biases. 

\section{Limitations}

Our study requires access to 
multiple LLMs to generate model performance vectors for each prompt in a benchmark.
This can be computationally expensive and require GPUs.
Some models, such as OpenAI's GPT-4, have limited API calls, making data collection time consuming. 

While we provide a novel approach
for researchers to investigate bias in their own studies, providing a comprehensive de-biasing methodology is not within the scope of this work. 

Finally, we have only touched the surface on why prompts have similar performance across multiple LLMs. 
There are many other components to investigate, such as the length of the prompt and prompt complexity.
This information could be leveraged to propose solutions on improving benchmarks, without running prompts through multiple LLMs.

\bibliography{llm_evaluation_custom, anthology}

\clearpage

\appendix

\section{Prompt structure}
\label{sec_app:prompts}

The prompts used for inference are depicted in Figures~\ref{fig:promptopenai},~\ref{fig:prompthellaswag},~\ref{fig:promptcommonsense} and \ref{fig:promptsumm} for ANLI, HellaSwag,  CommonsenseQA and CNN/Daily respectively.

\begin{figure}[h!]
  \centering
  \begin{tikzpicture}
   \node[draw, rounded corners=5pt, inner sep=2pt, text width=0.45\textwidth, align=center] (mysquare) at (0,0) {Given the following context:  \{premise\} \\ Question:\{hypothesis\} \\ True, False or Neither? \\ The answer is: };
  \end{tikzpicture}
  \caption{Prompt used during inference for ANLI.}
  \label{fig:promptopenai}
\end{figure}

\begin{figure}[h!]
  \centering
  \begin{tikzpicture}
   \node[draw, rounded corners=5pt, inner sep=2pt, text width=0.45\textwidth, align=center] (mysquare) at (0,0) {\#\#\# System: You are an AI assistant. Provide a detailed answer so user do not need to search outside to understand the answer. \\
\#\#\# User: Category: \{activity\_label\} \\
Text: \{ctx\} \\
Completion options: \\
(1) \{choice\_1\}  \\
(2) \{choice\_2\}  \\
(3) \{choice\_3\}  \\
(4) \{choice\_4\} \\
\#\#\#  Assistant: The most likely text completion is:};
  \end{tikzpicture}
  \caption{Prompt used during inference for HellaSwag.}
  \label{fig:prompthellaswag}
\end{figure}

\begin{figure}[h!]
  \centering
  \begin{tikzpicture}
   \node[draw, rounded corners=5pt, inner sep=2pt, text width=0.45\textwidth, align=center] (mysquare) at (0,0) {Question: \{\{question\}\} \\
Answer options: \\
(A) \{\{choiceA\}\} \\
(B) \{\{choiceB\}\} \\
(C) \{\{choiceC\}\} \\
(D) \{\{choiceD\}\} \\
(E) \{\{choiceE\}\} \\
The answer is:};
  \end{tikzpicture}
  \caption{Prompt used during inference for CommonsenseQA.}
  \label{fig:promptcommonsense}
\end{figure}

\begin{figure}[h!]
  \centering
  \begin{tikzpicture}
   \node[draw, rounded corners=5pt, inner sep=2pt, text width=0.45\textwidth, align=center] (mysquare) at (0,0) {\#\#\# Article:\\
\{Text to summarize\}\\
\#\#\# Summary:};
  \end{tikzpicture}
  \caption{Prompt used during inference for CNN/Daily Mail.}
  \label{fig:promptsumm}
\end{figure}

\section{Results: Semantically similar prompts}
\label{sec_app:similar_prompts}

For the statistical tests in Section~\ref{subsec:semantic_similarity_method}, we describe a set of linear regression models being generated where each model contains the prompt pairs of a specific single prompt.
Here, we display semantically similar prompts from these models where the semantic similarity coefficient is high and significant in explaining the model performance dependent variable.

In Table~\ref{tab:anli_regression_reason_example}, the ANLI reason component demonstrates that the prompts are adversarial because the model is unable to perform simple math operations. 
In other words, the prompts elicit the same mathematical operation task.
For CNN/Daily Mail data, the prompts either refer to the economy or global warming as seen in Table~\ref{tab:dailycnn_regression_reason_example}.
This entails that the models' performance had similar capabilities in extracting text about these subjects.


\begin{table*}
    \caption{List of ANLI reasons having high semantic similarity with model performance. }
    \label{tab:anli_regression_reason_example}
    \centering
    \begin{tabular}{cp{0.85\linewidth}}
    \hline
        Reason & Text\\
    \hline
    1 & it says osaka beat williams  6-2, 6-4. So osaka lost 6 games total. The system may have missed this as it did not add up the losses from both sets\\
    2 & The 1972–73 California Golden Seals had a 13–55–10 record - so they lost about 4 times as many [55] as they won [13]; the model might not know math.\\
    3 & Although Shigeko Sasamori was interviewed about this event, it's uncertain if she witnessed it personally. I think the system is confused because of so many matching words.\\
    4 & It does not state whether she was rebound leader - although her points total was tied with another player - which might have confused the model.\\
    5 & his record is 6-5 not 5-5\\
    \hline
    \end{tabular}
\end{table*}

\begin{table*}
    \caption{List of Daily/CNN grounded truth summaries having high semantic similarity with model performance. }
    \label{tab:dailycnn_regression_reason_example}
    \centering
    \begin{tabular}{cp{0.85\linewidth}}
    \hline
        Label & Text\\
    \hline
    1 & Jeffrey Sachs : Raw capitalism is the economics of greed . Last year was the Earth's hottest year on record, he says.\\
    2 & Adam Sobel : California's steps against drought are a preview for rest of U.S. and world. Tying climate change to weather doesn't rest on single extreme event, Sobel says. The big picture should spur us to prepare for new climates by fixing infrastructure, he says.\\
    3 & India predicted to outpace China as as world's fastest-growing economy in next year. China's economy is slowing after over 25 years of breakneck growth. But experts say India simply can't size up against China 's raw economic might.\\
    4 & Bill Richardson : U.S announced plan to cut greenhouse gas emissions by 26 to 28 percent below 2005 levels by 2025. He says China, India, major corporations, cities among those already setting goals for cutting emissions. U.S. must lead in this effort.\\
    \hline
    \end{tabular}
\end{table*}

\section{Results: Increasing size of benchmark}
\label{sec_app:benchmark_size_increase}

\begin{figure*}[h!]
    \centering
    \subcaptionbox{ANLI}
    {\includegraphics[width=.45\textwidth]{figures/anli/performance_mmr_simplified.pdf}}
%
    \subcaptionbox{HellaSwag}
    {\includegraphics[width=.45\textwidth]{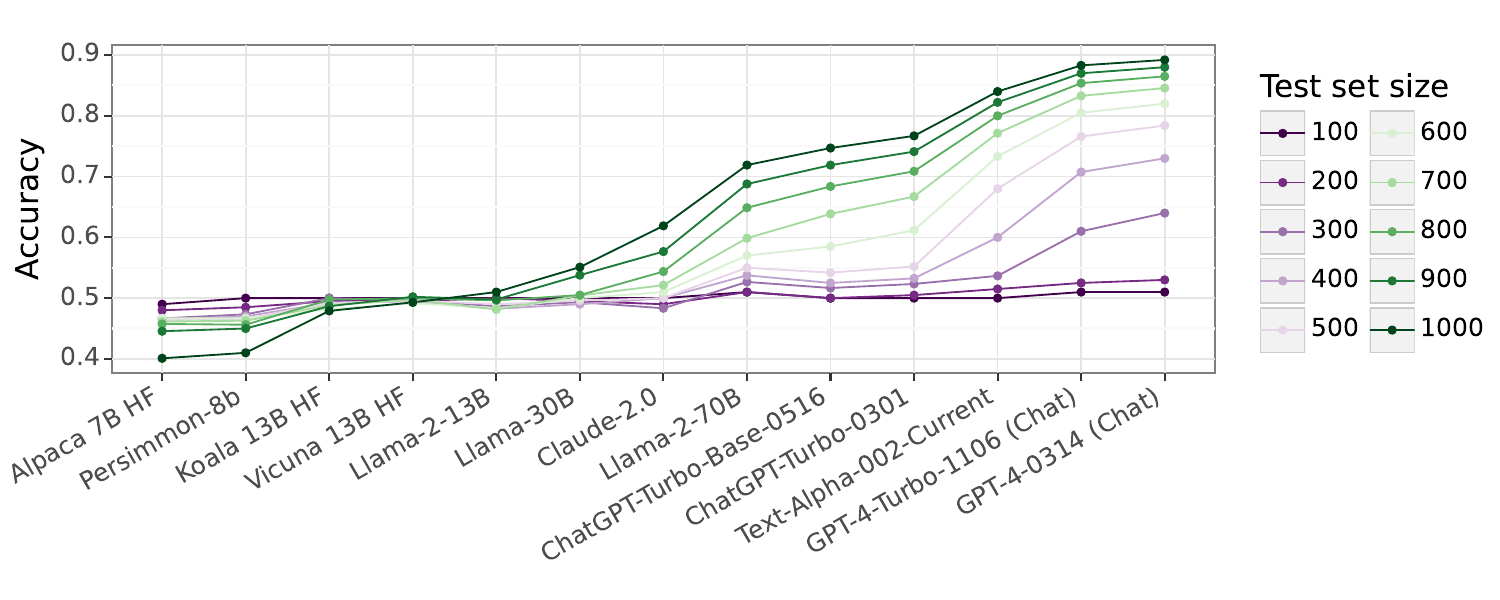}}
%
    \subcaptionbox{CommonsenseQA}
    {\includegraphics[width=.45
\textwidth]{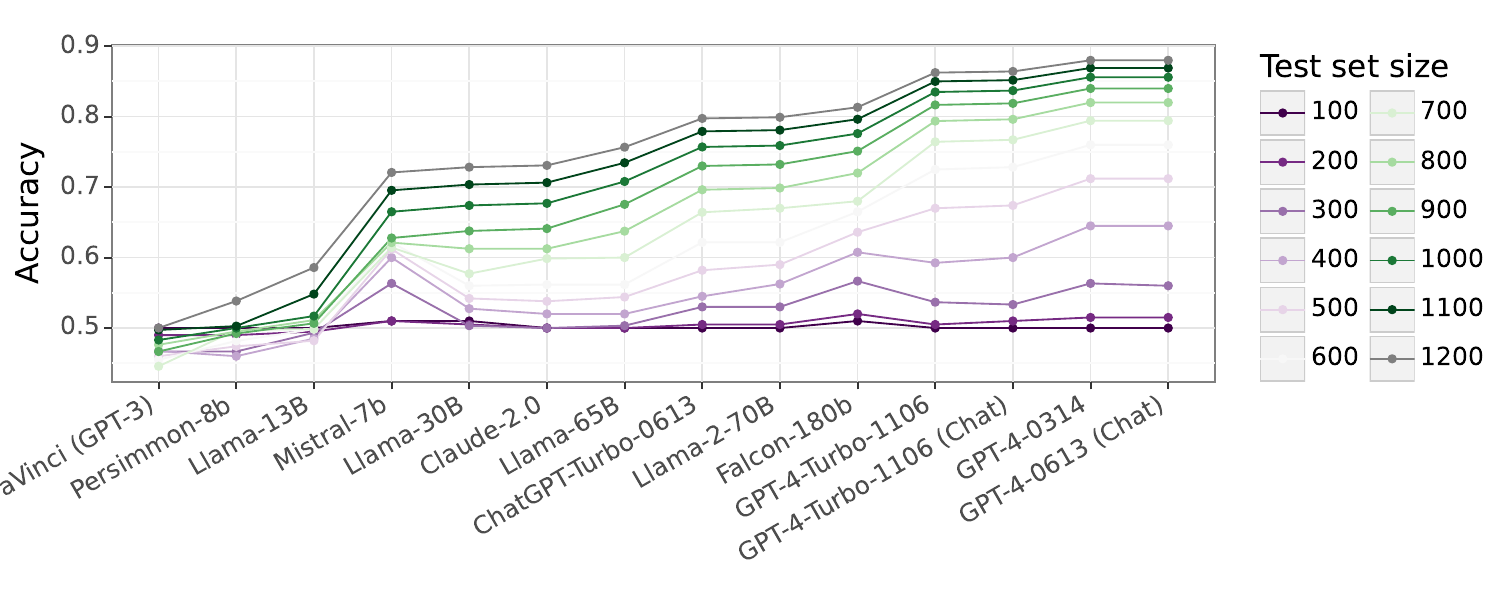}}
%
    \subcaptionbox{CNN/Daily Mail}
    {\includegraphics[width=.45\textwidth]{figures/cnn_daily_mail/performance_mmr_simplified.pdf}}
\caption{Average performance as benchmark size increases. Prompts are added to maximize average cosine distance. Maximum benchmark size corresponds to performance on the original benchmark.}
\label{fig:performance_mmr_appendix}
\end{figure*}

\begin{figure*}[h!]
    \centering
    \subcaptionbox{ANLI}
    {\includegraphics[width=.45\textwidth]{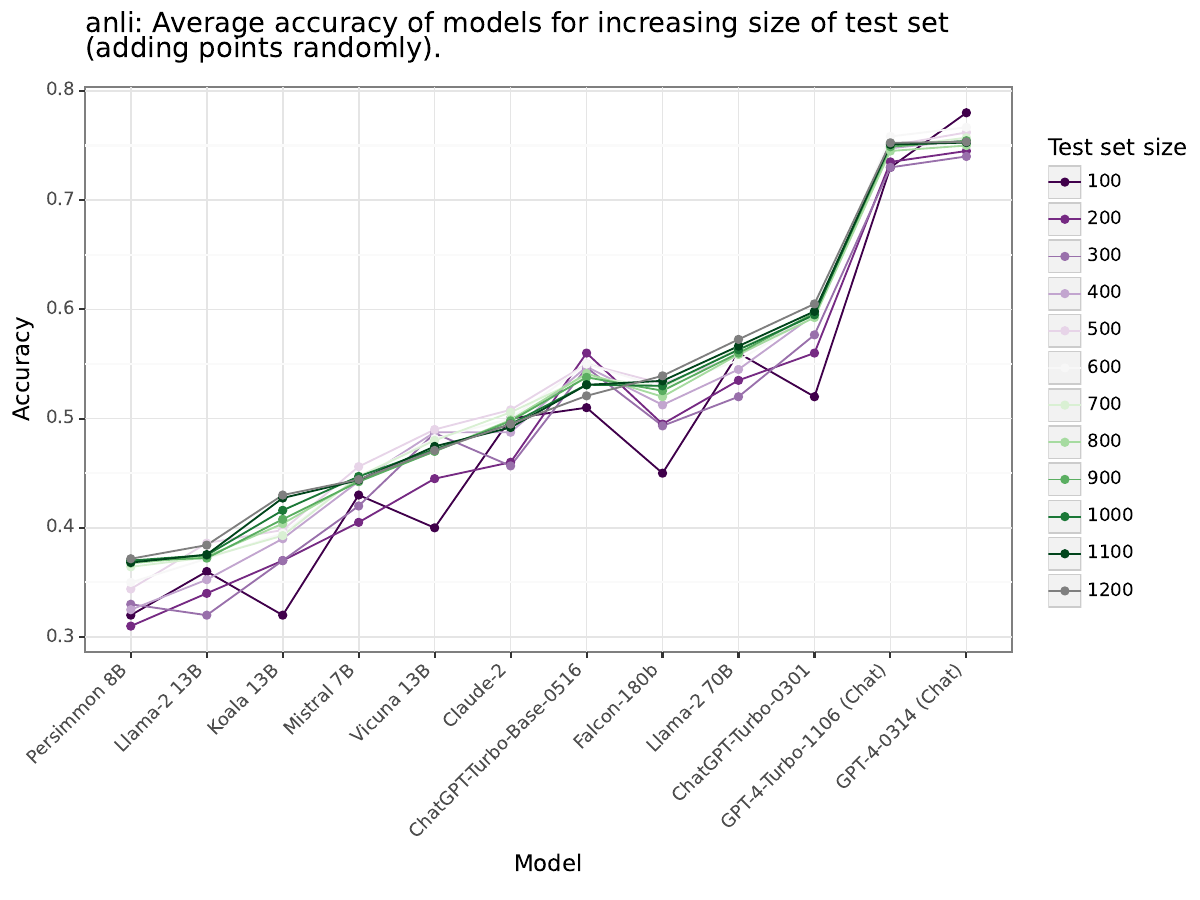}}
%
    \subcaptionbox{HellaSwag}
    {\includegraphics[width=.45\textwidth]{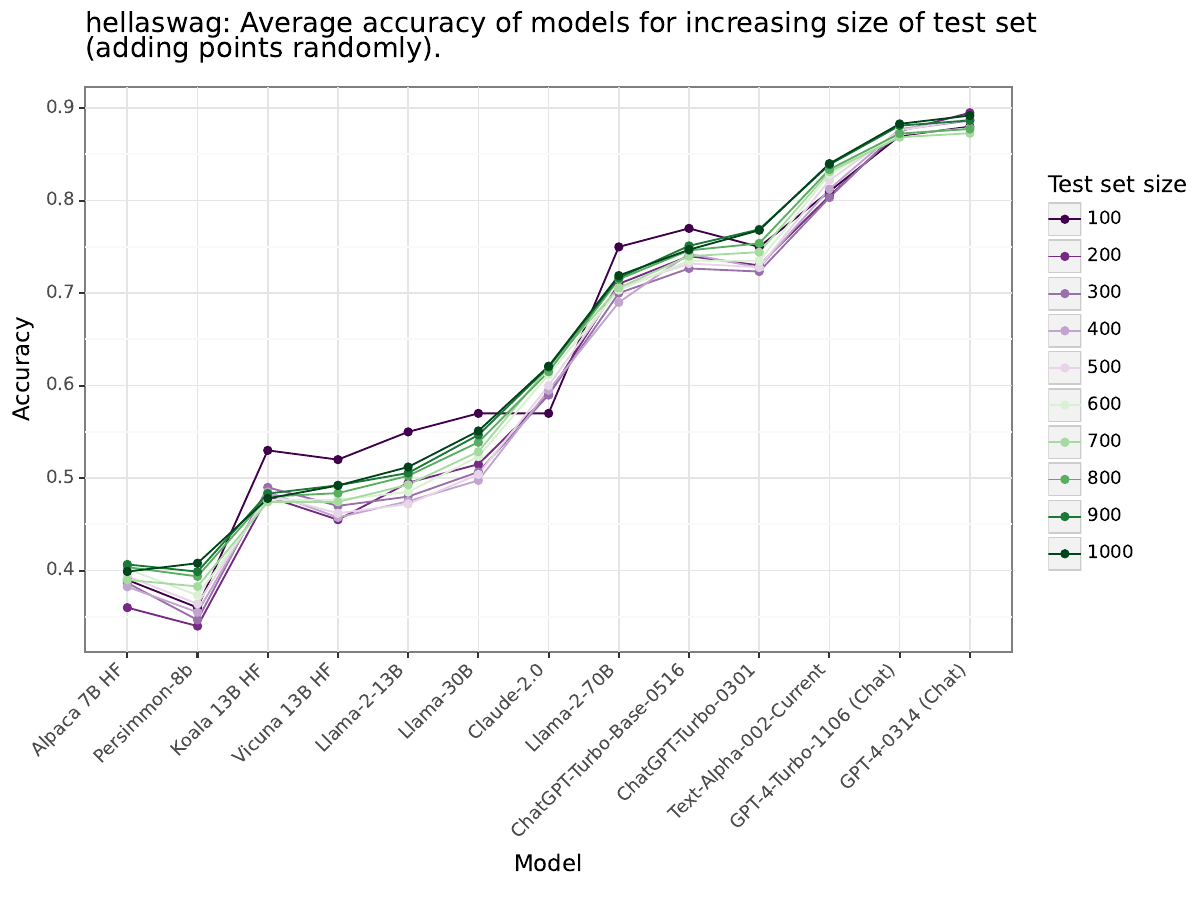}}
%
    \subcaptionbox{CommonsenseQA}
    {\includegraphics[width=.45\textwidth]{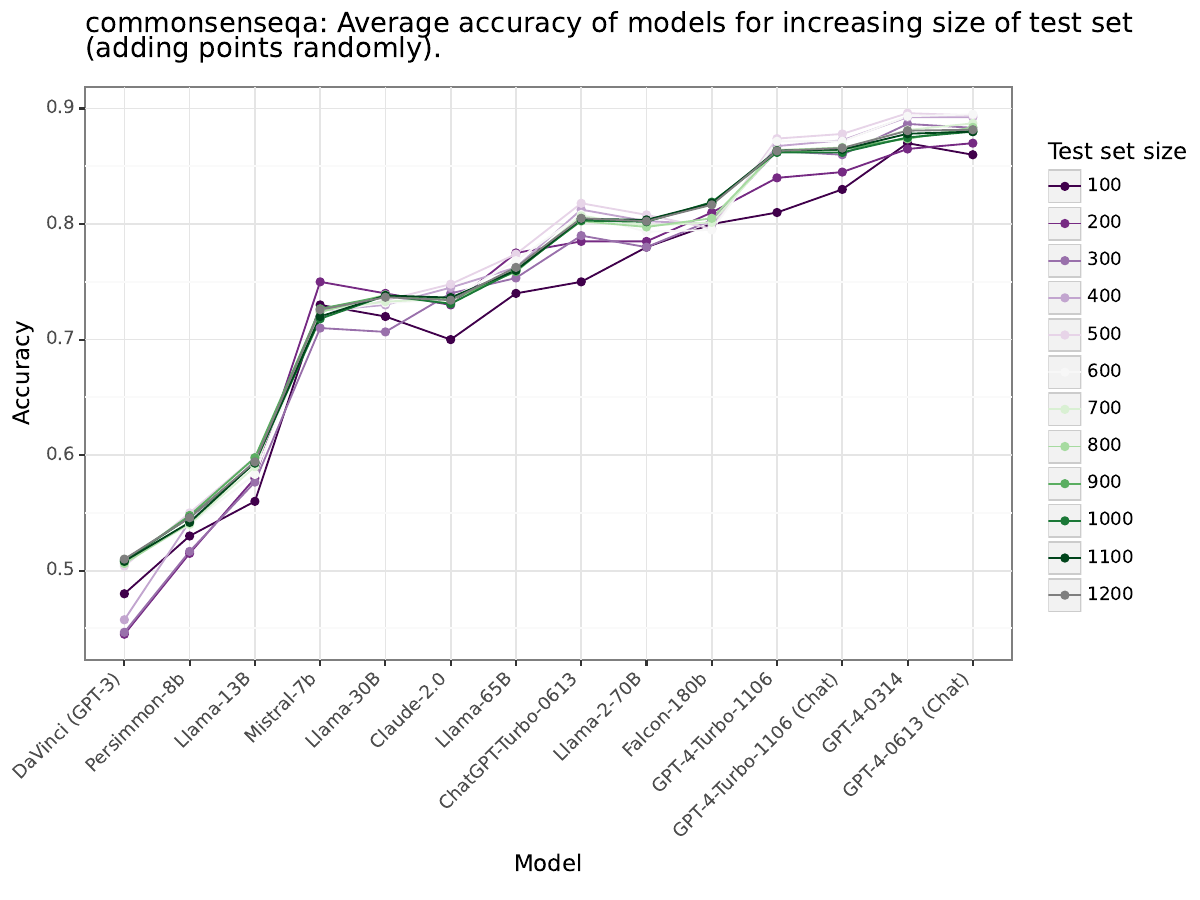}}
%
    \subcaptionbox{CNN/Daily Mail}
    {\includegraphics[width=.45\textwidth]{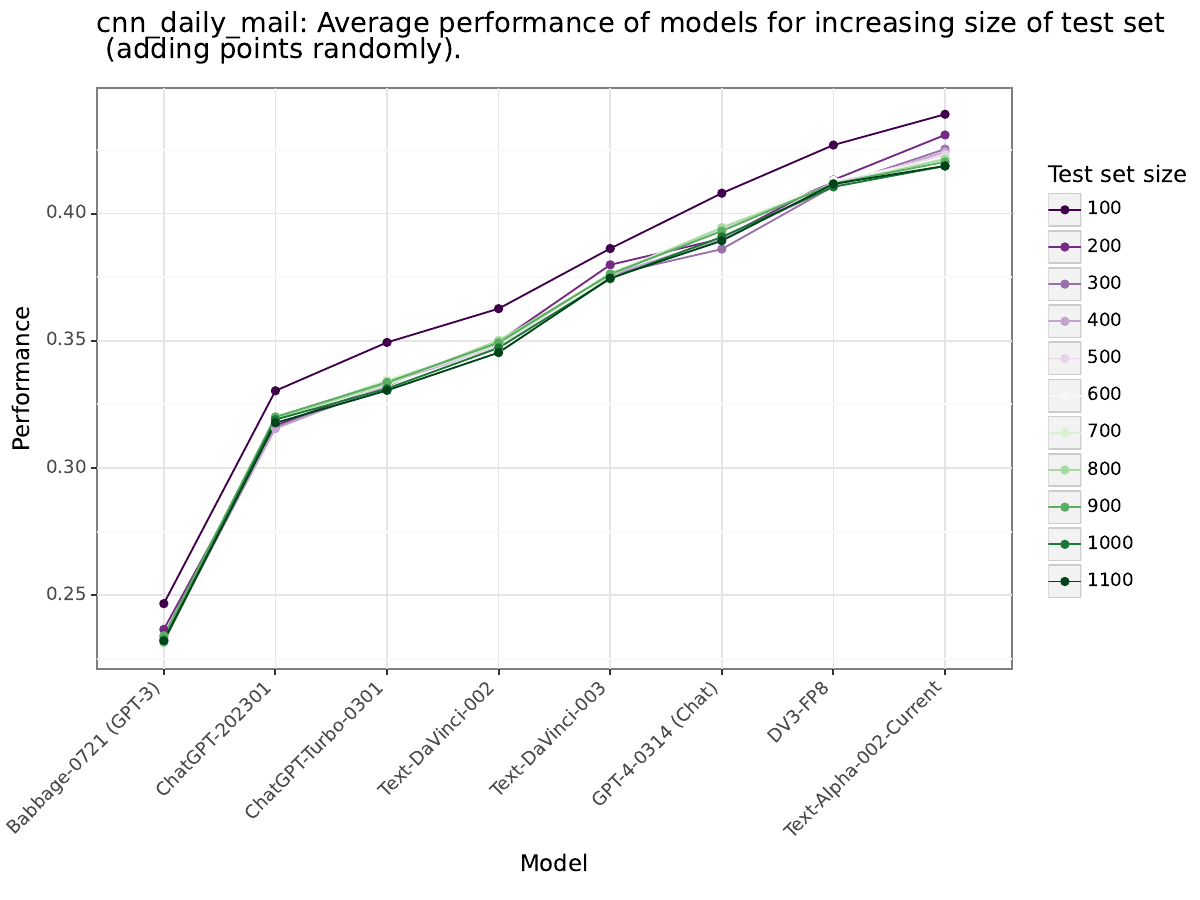}}
\caption{Average performance as benchmark size increases. Prompts are added in random order. Maximum benchmark size corresponds to performance on the original benchmark.}
\label{fig:performance_random_appendix}
\end{figure*}

Figure~\ref{fig:performance_mmr_appendix} shows results for all benchmarks for our experiments on increasing size of benchmark using the most informative method, as described in Section~\ref{subsec:increasing_size_benchmark}.
Figure~\ref{fig:performance_random_appendix} shows results for all benchmarks when adding prompts in random order. 

\section{Results: Distributions of weighted performance}
\label{sec_app:random_weights_plots}

\begin{figure*}[!h]
    \centering
    \subcaptionbox{\label{subfig:random_weights_violin_appendix} ANLI}
    {\includegraphics[width=.4\textwidth]{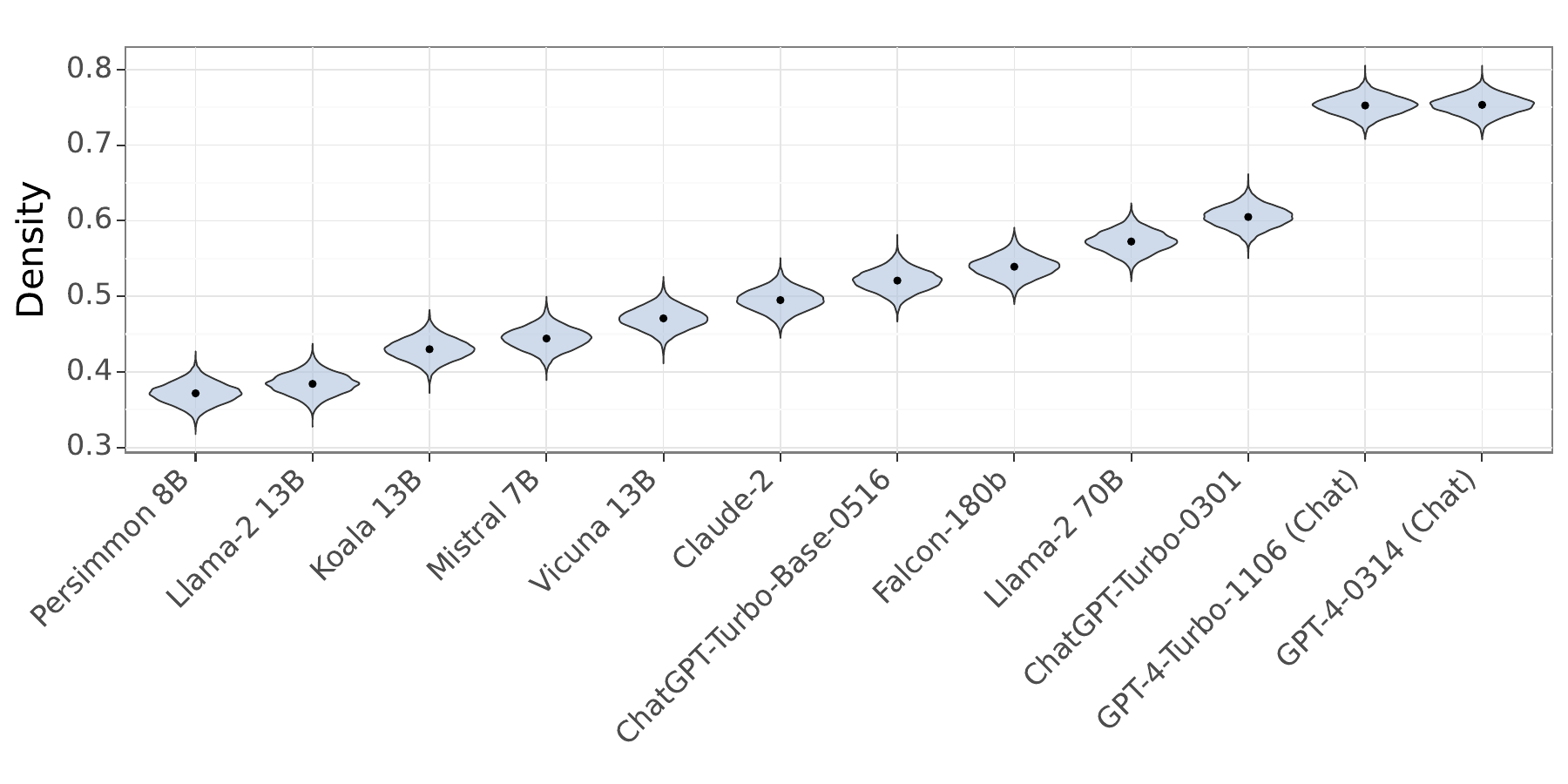}}
    \subcaptionbox{\label{subfig:random_weights_confusion_matrix_appendix} ANLI}
    {\includegraphics[width=.45\textwidth]{figures/anli/uniform_sampling_heatmap_simplified.pdf}}
    \subcaptionbox{\label{subfig:cnn_random_weights_violin_appendix} CNN/Daily Mail}
    {\includegraphics[width=.4\textwidth]{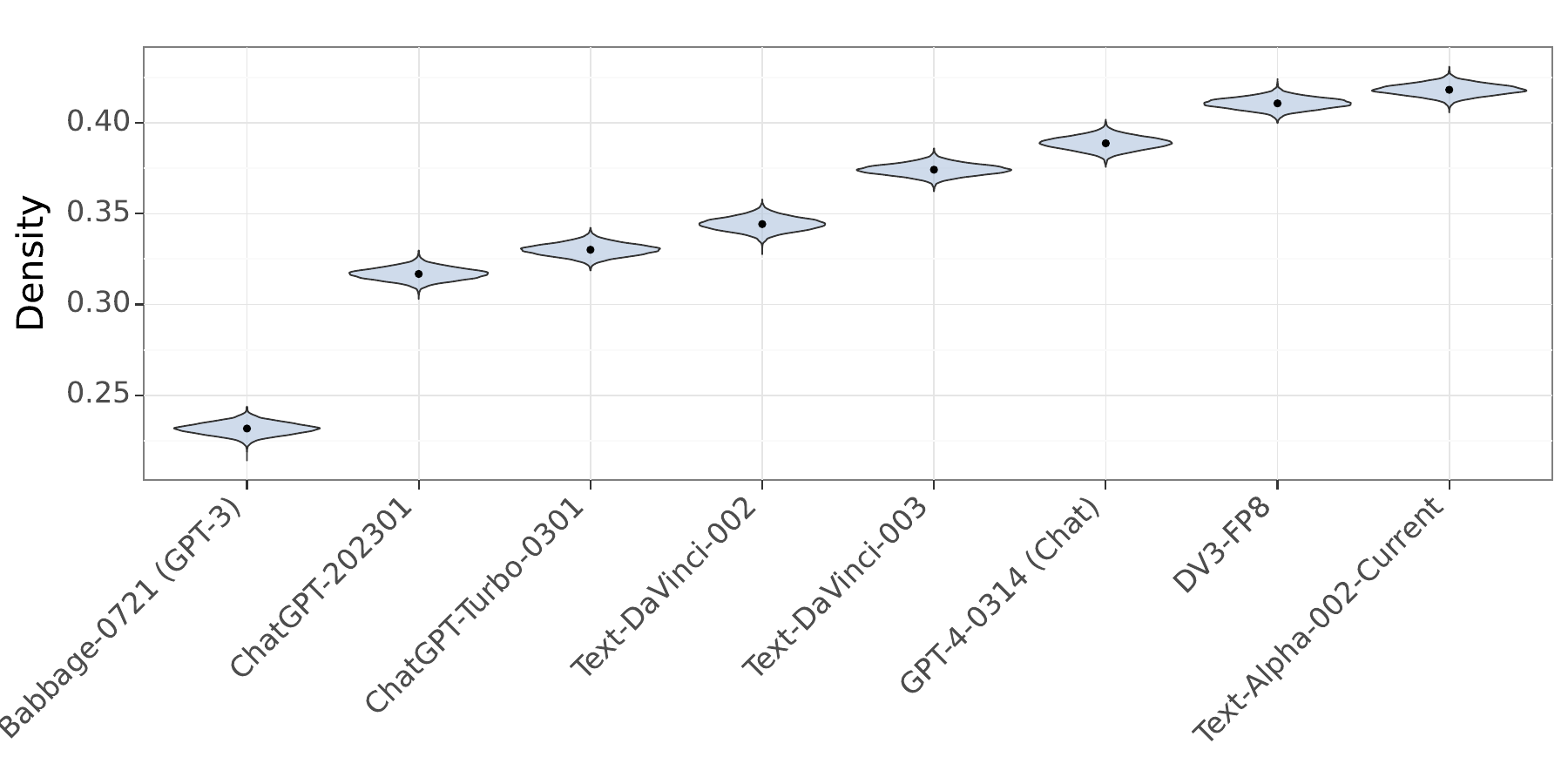}}
    \subcaptionbox{\label{subfig:cnn_random_weights_confusion_matrix_appendix} CNN/Daily Mail}
    {\includegraphics[width=.45\textwidth]{figures/cnn_daily_mail/uniform_sampling_heatmap_simplified.pdf}}
    \subcaptionbox{\label{subfig:hellaswag_random_weights_violin_appendix} HellaSwag}
    {\includegraphics[width=.4\textwidth]{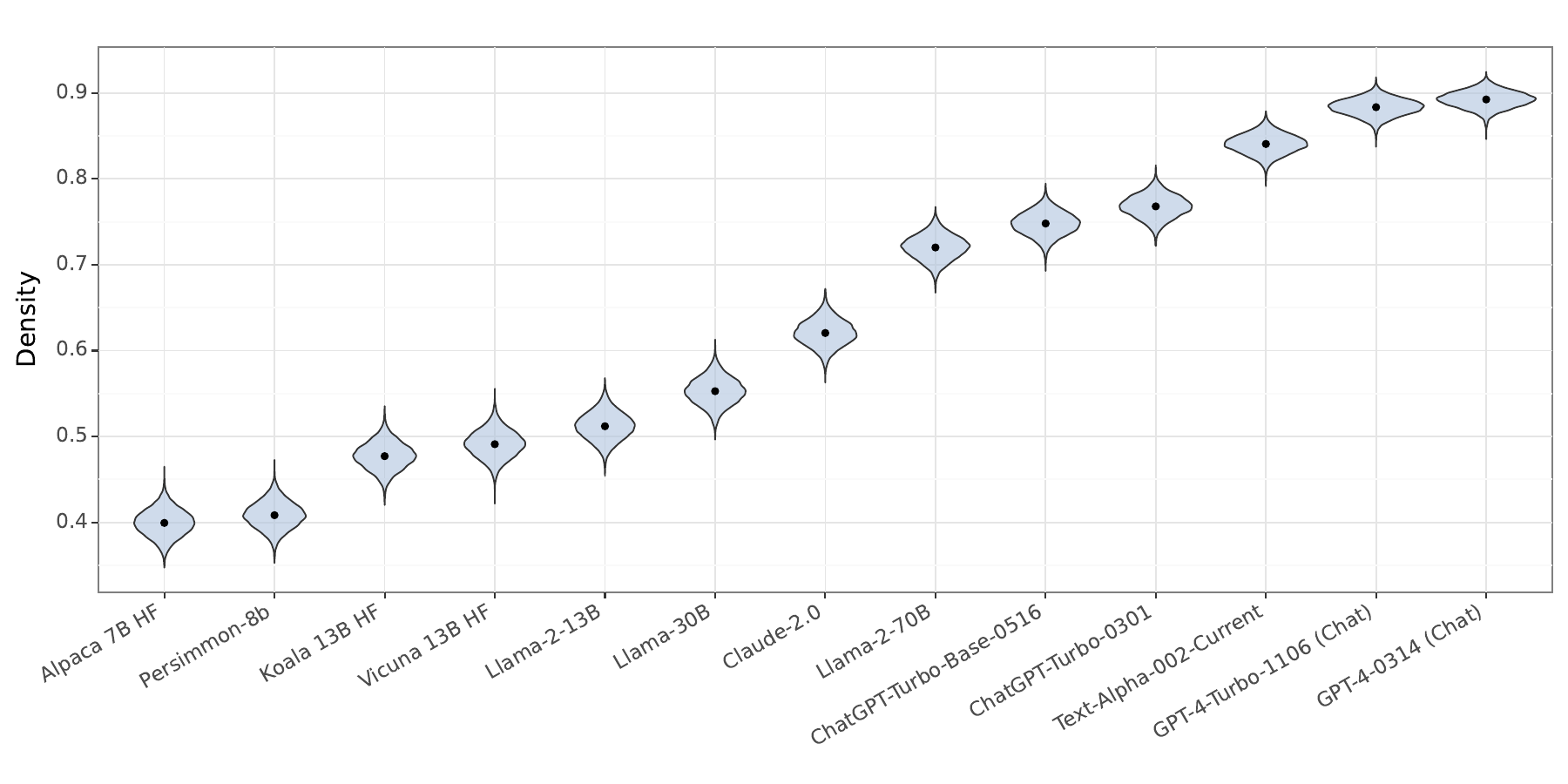}}
    \subcaptionbox{\label{subfig:hellaswag_random_weights_confusion_matrix_appendix} HellaSwag}
    {\includegraphics[width=.45\textwidth]{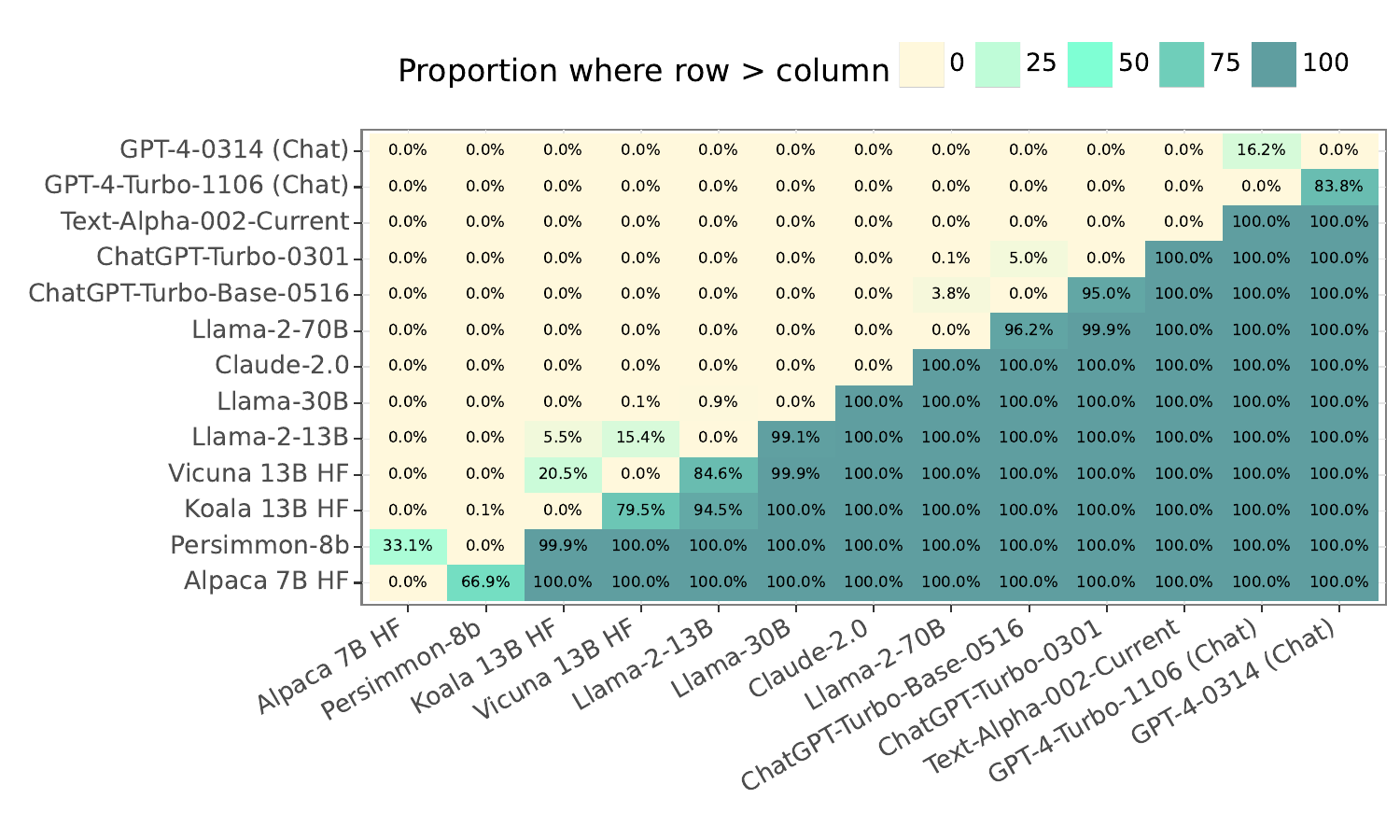}}
    \subcaptionbox{\label{subfig:CommonsenseQA_random_weights_violin_appendix} CommonsenseQA}
    {\includegraphics[width=.4\textwidth]{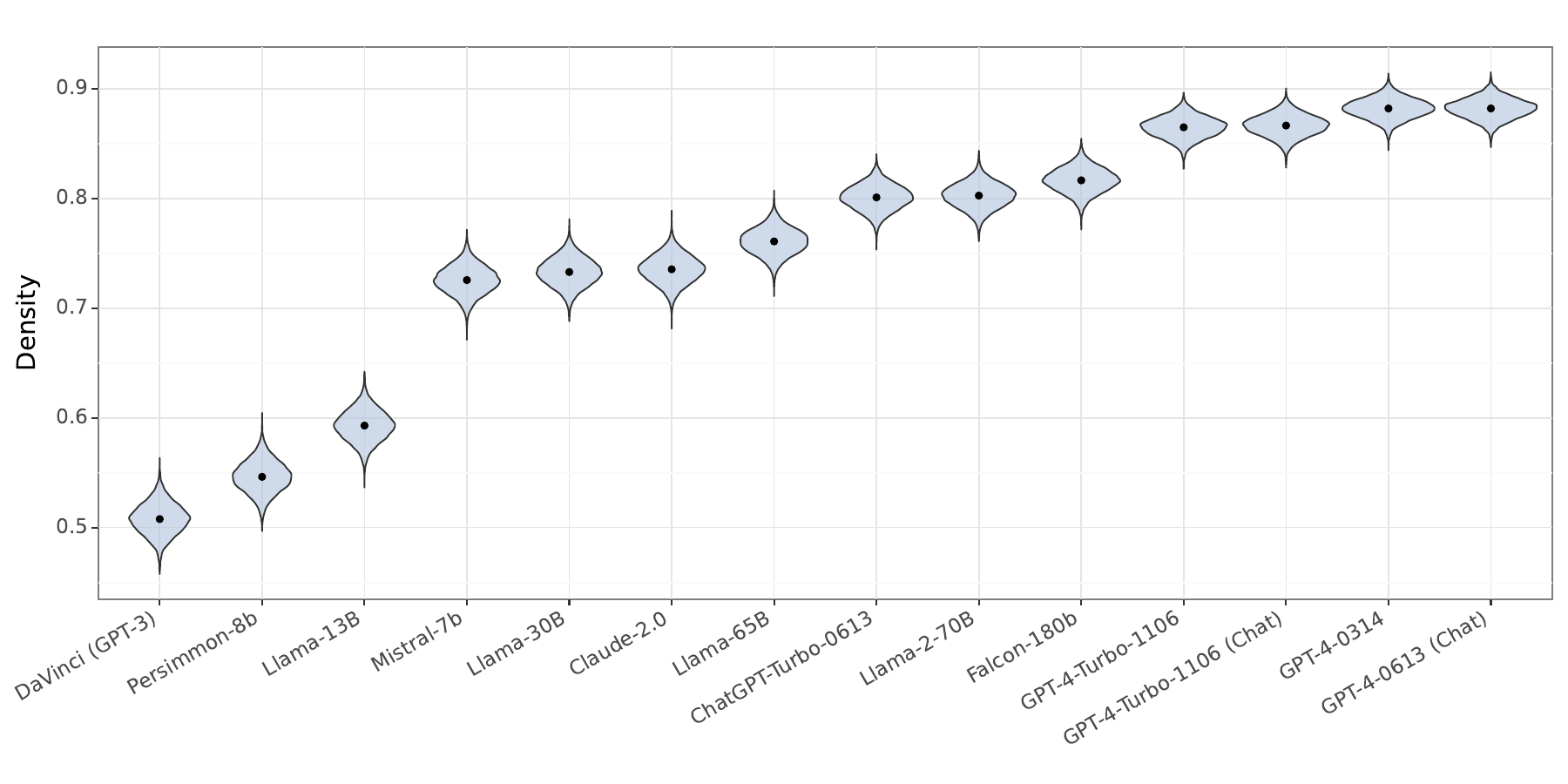}}
    \subcaptionbox{\label{subfig:CommonsenseQA_random_weights_confusion_matrix_appendix} CommonsenseQA}
    {\includegraphics[width=.45\textwidth]{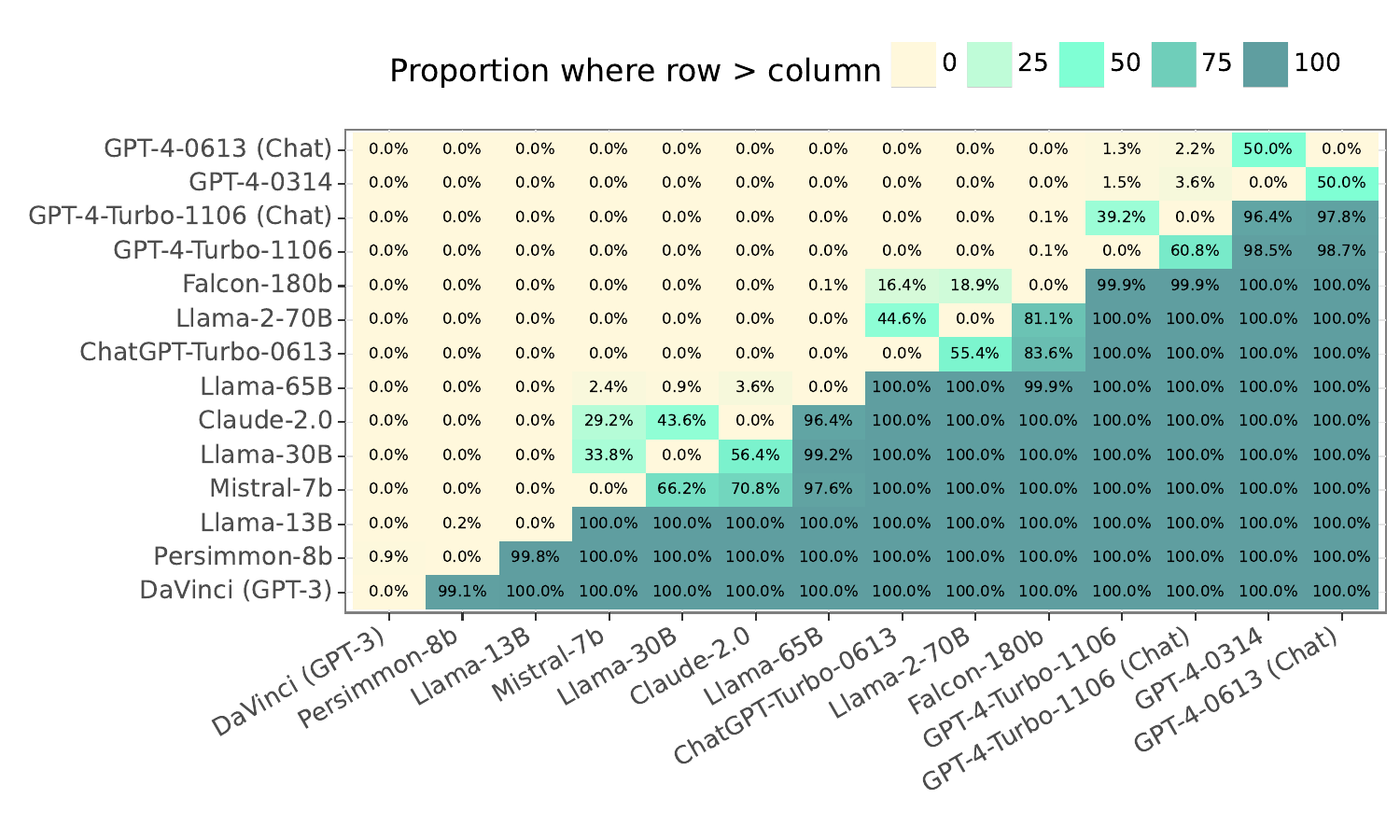}}
\caption{Left column: Distribution of weighted performance for randomly sampled weights. The black dot corresponds to performance when using uniform weights. 
Right column: Pairwise comparison of weighted performance. Every cell corresponds to the proportion of times the model in the row outperforms the model of the corresponding column.}
\label{fig:random_weights_all_bencharks}
\end{figure*}

Figure~\ref{fig:random_weights_all_bencharks} shows distribution of weighted performance and pairwise ranking changes for all benchmarks.

\end{document}